\documentclass{article}

\usepackage{arxiv}
\usepackage[utf8]{inputenc} % allow utf-8 input
\usepackage[T1]{fontenc}
\usepackage{color}
\usepackage{colortbl}
\usepackage[dvipsnames]{xcolor}
\definecolor{mgray}{gray}{.9}
\usepackage{wrapfig}
\usepackage{natbib}
\usepackage[pagebackref=true,colorlinks=true,citecolor=brown]{hyperref}
\renewcommand{\paragraph}[1]{\vspace{.1em}\noindent\textbf{#1}}
\usepackage[font=small]{caption}

\usepackage[utf8]{inputenc} % allow utf-8 input
\usepackage[T1]{fontenc}    % use 8-bit T1 fonts
\usepackage{hyperref}       % hyperlinks

\usepackage{url}            % simple URL typesetting
\usepackage{booktabs}       % professional-quality tables
\usepackage{amsfonts}       % blackboard math symbols
\usepackage{nicefrac}       % compact symbols for 1/2, etc.
\usepackage{microtype}      % microtypography
\usepackage{lipsum}		% Can be removed after putting your text content
\usepackage{graphicx}
\usepackage{doi}
\usepackage{caption}  % 提供 \captionof
\usepackage{textcomp,booktabs}
\usepackage{multirow}
\usepackage{amsmath}
\usepackage{enumitem}

\usepackage{booktabs, multirow, tabularx, siunitx}
\usepackage{svg}

\usepackage{capt-of}

\usepackage{adjustbox}

\usepackage{subcaption}
\usepackage{placeins}
\usepackage{pifont}
\newcommand{\xmark}{\ding{55}}

\makeatletter
\newcommand{\labelline}[1]{%
  \begingroup
    \edef\@currentlabel{\number\value{ALG@line}}%
    \label{#1}%
  \endgroup
}

\makeatother

\usepackage{algorithm}
\usepackage{algpseudocode}
\algrenewcommand\algorithmicrequire{\textbf{Input:}}
\algrenewcommand\algorithmicensure{\textbf{Output:}}
\algnewcommand{\LineComment}[1]{\State \(\triangleright\) #1}

\title{Towards Safe Mobility: A Unified Transportation Foundation Model enabled by  Open-Ended Vision–Language Dataset}

\author{
Wenhui~Huang$^{1,2}$ \qquad \qquad \qquad
\And
Songyan~Zhang\footnotemark[3] \ $^1$ \qquad \qquad \qquad
\And
Collister~Chua\footnotemark[3] \ $^1$ \qquad \qquad
\And 
Yang~Liang$^{1}$ \qquad \qquad \qquad
\And
Zhiqi~Mao$^{1}$ \qquad \qquad
\And
Heng~Yang$^{2}$ \qquad \qquad
\And 
Chen~Lv\footnotemark[2] \ $^1$
}

\hypersetup{
pdftitle={A template for the arxiv style},
pdfsubject={q-bio.NC, q-bio.QM},
pdfauthor={David S.~Hippocampus, Elias D.~Striatum},
pdfkeywords={First keyword, Second keyword, More},
}

\begin{document}
\maketitle
\renewcommand{\thefootnote}{}
\footnotetext{$^\ddagger$ Project Lead. $^\dagger$ Corresponding Author.}
\renewcommand{\thefootnote}{\arabic{footnote}}

\renewcommand{\thefootnote}{}
\footnotetext{$^1$ Nanyang Technological University. $^2$ Harvard University.}
\renewcommand{\thefootnote}{\arabic{footnote}}

\vspace{-10mm}

\begin{center}
    % \url{https://github.com/OscarHuangWind/UniVLT.git}
    \href{https://github.com/OscarHuangWind/UniVLT.git}{Source Code, Dataset, Models}
\end{center}

\vspace{5mm}

\begin{abstract}
Urban transportation systems face growing safety challenges that require scalable intelligence for emerging smart mobility infrastructures. While recent advances in foundation models and large-scale multimodal datasets have strengthened perception and reasoning in intelligent transportation systems (ITS), existing research remains largely centered on microscopic autonomous driving (AD), with limited attention to city-scale traffic analysis. In particular, open-ended safety-oriented visual question answering (VQA) and corresponding foundation models for reasoning over heterogeneous roadside camera observations remain underexplored. To address this gap, we introduce the Land Transportation Dataset (LTD), a large-scale open-source vision--language benchmark for open-ended reasoning in urban traffic environments. LTD contains 11.6K high-quality VQA pairs collected from heterogeneous roadside cameras, spanning diverse road geometries, traffic participants, illumination conditions, and adverse weather. The dataset integrates three complementary tasks: fine-grained multi-object grounding, multi-image camera selection, and multi-image risk analysis, requiring joint reasoning over minimally correlated views to infer hazardous objects, contributing factors, and risky road directions. To ensure annotation fidelity, we combine multi-model vision--language generation with systematic cross-validation and human-in-the-loop refinement. Building upon LTD, we further propose UniVLT, a transportation foundation model trained via curriculum-based knowledge transfer to unify microscopic AD reasoning and macroscopic traffic analysis within a single architecture. Extensive experiments on LTD and multiple AD benchmarks demonstrate that UniVLT achieves state-of-the-art performance on open-ended reasoning tasks across diverse domains, while exposing limitations of existing foundation models in complex multi-view traffic scenarios. We release the dataset and code to facilitate reliable multimodal research for intelligent transportation systems.
\end{abstract}

% keywords can be removed
% \keywords{First keyword \and Second keyword \and More}

\section{Introduction}
\label{sec1}

\begin{figure*}[t]
  \centering
  \includegraphics[width=\textwidth]{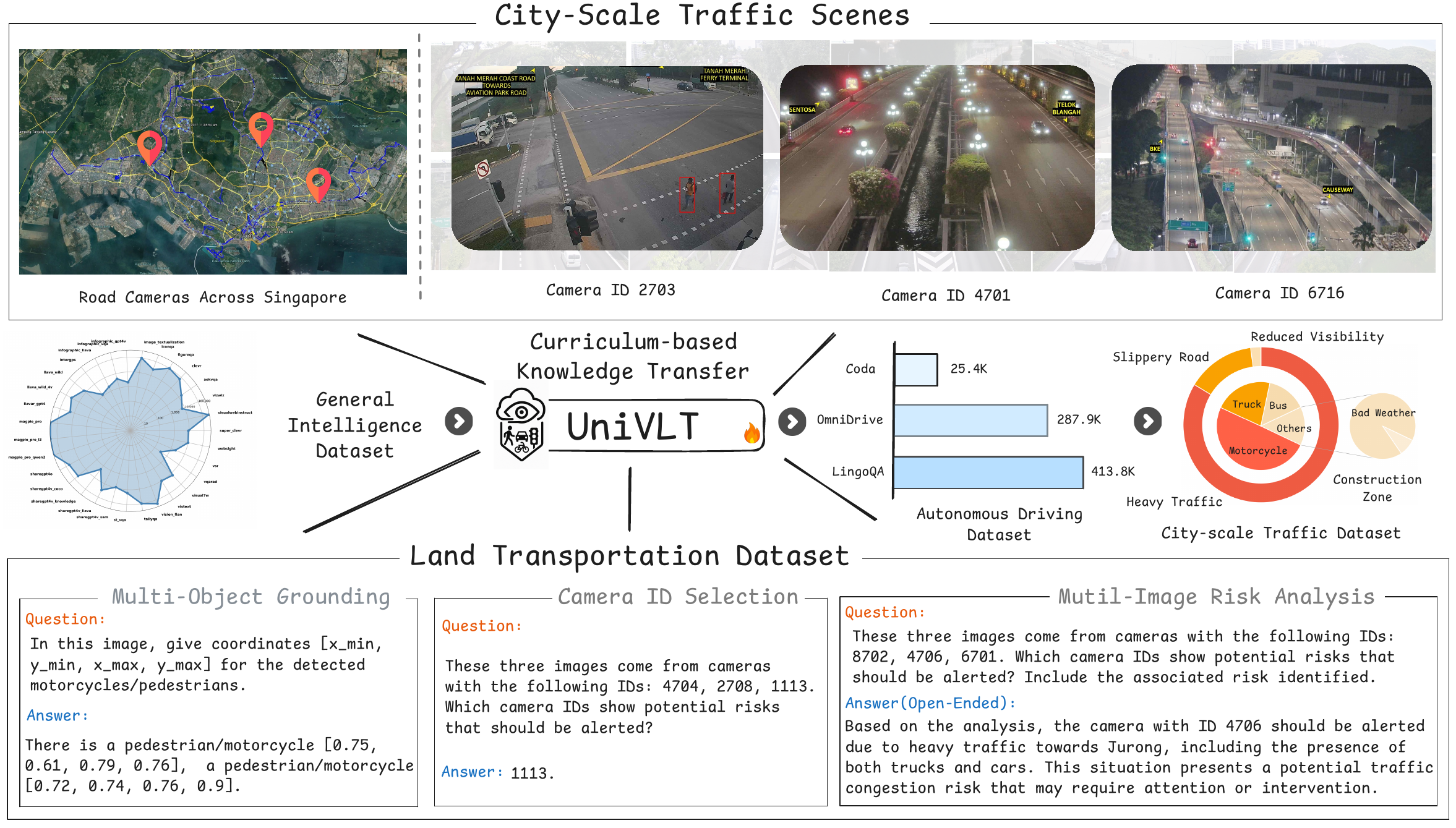}
  % \vspace{-10pt}
  \caption{Overview of the land transportation dataset (LTD) and unified vision-language-transportation Model (UniVLT). LTD consists of three types of reasoning tasks: fine-grained multi-object grounding, multi-image camera id selection, and open-ended multi-image risk analysis. UniVLT is trained via curriculum-based knowledge transfer, progressively transitioning from general-domain learning to autonomous driving (AD) and further to the traffic domain. Through this training strategy, UniVLT is capable of handling multiple objects and reasoning over multiple images captured by roadside cameras across diverse road segments, even when the visual inputs exhibit minimal or no explicit inter-relationships.}
  \label{fig:teaser}
  % \vspace{-25pt}
\end{figure*}

Ensuring road safety and efficient mobility remain central objectives of modern Intelligent Transportation Systems (ITS). With the rapid growth of urban populations and travel demand, transportation networks are becoming increasingly complex and congested, resulting in higher risks of traffic incidents and inefficiencies in traffic management ~\citep{its, elassy_its}. Urban traffic environments involve dynamic interactions among heterogeneous road users, including pedestrians, cyclists, and motorcyclists, across multiple spatial and temporal scales. Addressing safety risks and improving traffic efficiency therefore requires intelligent systems capable of perceiving, understanding, and reasoning about complex traffic situations. Recent advances in data-driven technologies and large-scale sensing infrastructure, such as roadside cameras and connected sensing platforms, have created unprecedented opportunities to develop intelligent analytical tools that support safer and more responsive transportation systems ~\citep{zado_its_sensor}.

Recent advances in foundation models, including large language models (LLMs) and vision-language models (VLMs), have demonstrated significant potential for enhancing scene understanding and analytical capabilities within ITS. Broadly speaking, ITS research can be examined from two complementary perspectives: a microscopic level, which focuses on individual vehicle perception and decision-making in autonomous driving (AD), and a macroscopic level, which concerns large-scale traffic monitoring, planning, and management across urban transportation networks. By leveraging large-scale multi-modal pretraining and commonsense reasoning capabilities exhibited by foundation models, new opportunities emerge to enhance mobility safety across multiple domains of transportation intelligence, ultimately contributing to improved traffic safety~\citep{trafficvlm}. 

At the microscopic level, previous studies on foundation models within ITS have predominantly emphasized improving interpretability and multitask reasoning capabilities from the perspective of autonomous vehicles (AVs). Most approaches adopt a question answering (QA) paradigm to unify diverse driving-related tasks, including object recognition, visual grounding, scene understanding, risk assessment, and strategic driving decisions~\citep{openread, alphamayo, mao2023gpt}.
This paradigm is supported by open-source AD datasets such as LingoQA~\citep{lingoqa}, DRAMA~\citep{drama}, and CODA~\citep{coda}, which enable targeted training for handling complex social interactions and rare-event cases, often referred to as curse-of-rarity scenarios~\citep{cor}. Recent developments further extend these VLM paradigms toward direct action planning tasks. Such Vision-Language-Action (VLA) models \citep{vla} bridge perception, reasoning, and vehicle control within an integrated framework, thus enabling end-to-end autonomous driving solutions. Experimental results in simulated environments, notably the CARLA simulator, as well as in real-world driving conditions, confirm the viability and potential of these approaches for autonomous driving tasks \citep{lmdrive, drivevlm, autovla}.

In contrast to the flourishing advancements observed within the AD domain~\citep{hu2025vision}, relatively few studies have explored the potential benefits of foundation models from a macroscopic transportation perspective. As urban environments globally face increasingly intricate transportation challenges, there is an urgent need for intelligent, scalable, and proactive solutions to enhance overall road safety. Traditional methodologies, including traffic flow analysis, public safety campaigns, and manual surveillance, serve as essential components for urban transportation management yet are proving increasingly inadequate when confronted with the complexity and evolving nature of contemporary mobility issues.

A prominent example is the persistent road safety concerns in Singapore, exacerbated by changing mobility patterns and demographic shifts, particularly the aging population~\citep{sg_ageing}. Elderly pedestrians and motorcyclists, identified as highly vulnerable road-user groups, continue to be disproportionately represented in traffic accident and fatality statistics~\citep{vrus}. Despite a 6.9\% reduction in elderly pedestrian injuries from 2023 to 2024, this demographic, which constitutes merely 12.5\% of the Singaporean population, remains significantly overrepresented, accounting for nearly 44\% of pedestrian fatalities. Additionally, motorcyclists, although a relatively small portion of the overall vehicle population, experienced a notable 25\% increase in accident involvement in 2024. Concurrently, motorcycle usage has incrementally increased from 14.4\% in 2023 to 14.7\% in 2024, further elevating associated safety risks \citep{jordan2015machine}. Singapore's situation reflects broader global trends, as numerous metropolitan areas confront similar transportation safety and management challenges \citep{duan2024cityllava}. Consequently, a reconsideration of prevailing strategies for managing transportation safety and efficiency is critically necessary. Leveraging foundation models offers significant potential, not only for the real-time detection and mitigation of road incidents but also in alleviating substantial human resource demands typically associated with post-incident analysis and traffic measurement operations. Harnessing these advanced models could fundamentally transform how urban safety and traffic efficiency are managed, facilitating more responsive and sustainable urban transportation systems.

To address this research gap, we introduce a city-scale traffic dataset, termed the Land Transportation Dataset (LTD), together with an associated unified vision--language--transportation (UniVLT) model. LTD comprises a diverse collection of images captured by roadside cameras across Singapore and is annotated for three core tasks: fine-grained multi-object grounding, multi-image camera ID selection, and open-ended multi-image risk analysis. To further enhance dataset quality, we adopt a human-in-the-loop (HITL) annotation paradigm to revise counterfactual and hallucinated annotations, and employ a top-$k$ answer selection strategy to promote diversity in open-ended reasoning annotations. To the best of our knowledge, LTD introduces the first open-ended traffic VQA dataset at city scale based on roadside camera imagery, moving beyond closed-form answer spaces and explicitly challenging VLMs to reason through complex traffic scenarios without relying on elimination strategies or dataset biases. Additionally, we introduce a transportation-tailored VLM, termed UniVLT, trained through multi-stage fine-tuning. As a result, UniVLT not only supports multi-task learning from the microscopic AV perspective, but also extends open-ended reasoning to macroscopic traffic scales, unifying knowledge from both AD and traffic domains within a single model. Figure~\ref{fig:teaser} illustrates the overview of the proposed LTD and UniVLT. The primary contributions of this work are summarized as follows:

\begin{enumerate}

\item We present a systematic AI framework for smart mobility, spanning large-scale vision–language dataset construction, curation, and corresponding transportation foundation model development. This framework establishes a scalable pipeline for developing foundation-model-based solutions for city-scale intelligent transportation systems.

\item We introduce LTD, a city-scale traffic dataset collected from roadside cameras across Singapore and carefully annotated for three primary tasks: fine-grained multi-object grounding, multi-image camera ID selection, and open-ended multi-image risk analysis. To the best of our knowledge, LTD is the first open-ended traffic VQA dataset, bridging the gap between the growing demand for open-ended reasoning in intelligent transportation systems and the lack of corresponding VQA benchmarks.

\item We present a transportation-tailored VLM, termed UniVLT, fine-tuned through a curriculum-based knowledge transfer strategy that unifies microscopic autonomous driving and macroscopic traffic-domain knowledge within a single model.

\item The quality of the proposed LTD is rigorously validated by benchmarking UniVLT against a diverse set of pre-trained and domain-specific models.
In addition, we report comprehensive quantitative results of UniVLT across multiple real-world benchmarks, establishing a strong and reproducible baseline for future research in the community.

\end{enumerate}

The remainder of this paper is organized as follows: Sec. \ref{sec2} reviews the applications of foundation models in both AVs and ITS domains. Sec. \ref{sec3} elaborates the LTD and training strategy of UniVLT. Comprehensive benchmarks, discussions and ablation studies are presented in Sec. \ref{sec4} and \ref{ablation}. Finally, Sec. \ref{sec6} provides the conclusion, limitations, and future work.  
 
\section{Related Work}
\label{sec2}
\subsection{Vision-Language Models and Datasets for Autonomous Driving}
With the advent of vision–language models (VLMs), which augment vision encoders with large language models (LLMs), end-to-end (E2E) autonomous driving (AD) has shifted from a purely data-driven paradigm to a knowledge-driven one. Unlike conventional data-driven methods that rely on carefully curated datasets from a single domain, knowledge-driven AD leverages massive and diverse datasets across domains, enabling broader generalization. Distinct from prior works such as UniAD \citep{uniad}, VAD \citep{vad,vadv2}, and Interfuser \citep{interfuser}, these AD-tailored VLMs exploit artificial general intelligence (AGI), or common knowledge, so that multiple tasks can be unified within a single model while providing interpretability through natural language. For example, DriveLM \citep{drivelm} integrates VLMs trained on web-scale data into E2E driving systems, enhancing generalization through graph-based visual question answering (VQA) across perception, prediction, and planning. DriveMLM \citep{drivemlm} further incorporates traffic rules and high-level user instructions as inputs and produces semantic decisions with explanations that can be delivered to downstream behavioral planning. LMDrive \citep{lmdrive} pioneers training VLMs to directly output control signals and demonstrates large-scale closed-loop evaluation in the CARLA benchmark \citep{dosovitskiy2017carla}. DriveVLM \citep{drivevlm} takes a step forward by integrating VLMs into conventional AD pipelines and bringing the models from digital contexts into real-world deployment. More recently, RoboTron-Drive \citep{drivemm} addresses the challenge of heterogeneous modalities across open-source datasets. For instance, CODA-LM \citep{coda} provides only front-view images, LingoQA \citep{lingoqa} and DRAMA \citep{drama} provide multiple frames in sequence, and NuScenes-QA \citep{nuscenesqa} contains multi-camera views. To address abovementioned issue, RoboTron-Drive proposes an all-in-one VLM that efficiently leverages these diverse resources.

\subsection{Vision-Language Models and Datasets for Transportation}
In contrast to the flourishing datasets within the AD domain, relatively fewer macro-level VQA datasets, such as city-scale traffic datasets, are open-sourced. This limitation is likely due to sensitivity and security issues, which hinder the broader application of VLMs in intelligent transportation systems (ITS). The SUTD-TrafficQA~\citep{sutd-trafficqa} dataset provides 62.5k QA pairs covering six challenging tasks in multiple-choice format over complex real-world traffic scenes. Building on this, TRIVIA \citep{trivia} curates the dataset and trains VLMs with weak supervision for transportation reasoning. However, as described in their paper, the trained VLM is not quantitatively evaluated and its performance is affected by noisy automatic annotations. Subsequently, Woven by Toyota releases WTS \citep{wts}, a pedestrian-centric traffic video dataset for fine-grained spatio-temporal understanding, and benchmarks TrafficVLM \citep{trafficvlm} on it with multiple language metrics. Nevertheless, TrafficVLM is trained from scratch using only traffic datasets, with an individual vision encoder and T5-Base \citep{t5base} as the language decoder, and is specifically designed for its own setting, leading to a lack of generalization. More recently, TUMTraffic-VideoQA \citep{tumtraffic} proposes a new dataset collected from road cameras that comprises multiple-choice questions, object captioning, and object grounding annotations. However, this dataset is still under preparation for open-source release.

While acknowledging the progress of existing work, it is important to emphasize that the capability to reason over multiple images captured from different road cameras, where the images are not directly correlated, is critical for transportation VLMs. For example, an ideal model should be able to detect potential risks from several safety-critical camera views simultaneously and provide alerts and risk analyses to humans. Furthermore, unifying knowledge at both the micro level (AD) and macro level (traffic) is more beneficial than relying solely on the limited number of traffic datasets. To this end, the proposed UniVLT adopts a two-stage fine-tuning strategy: it first learns diverse knowledge from AD domain datasets, and then transfers to traffic-level VQA tasks to acquire the capability of handling multiple uncorrelated camera views, while retaining previously learned knowledge. We argue that such a unified VLM is crucial not only for traffic monitoring and risk alert systems but also for enabling future smart cities equipped with vehicle-to-vehicle (V2V) and vehicle-to-everything (V2X) communication systems.

\begin{figure*}[t]
    \centering
    \includegraphics[width=1.0\linewidth]{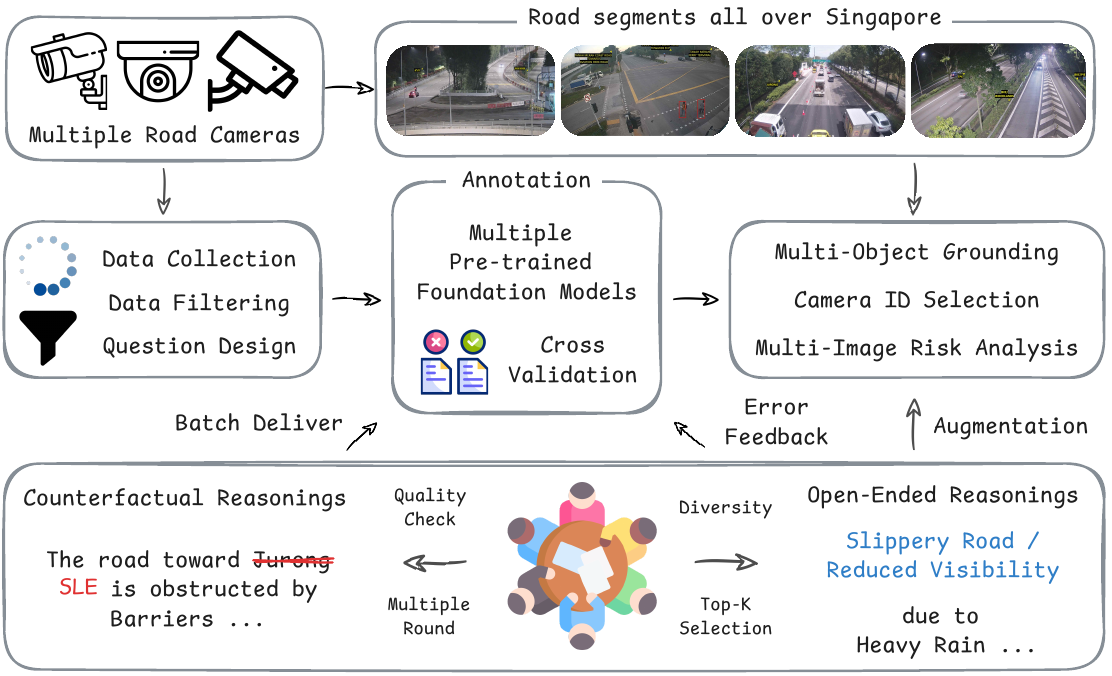}
    \caption{The pipeline of data annotation and curation with cross-validation of multiple pre-trained foundation models and human-in-the-loop qualification.}
    \vspace{-0.3cm}
    \label{datacuration}
\end{figure*}

\begin{figure*}[t]
    \centering
    \includegraphics[width=1.0\linewidth]{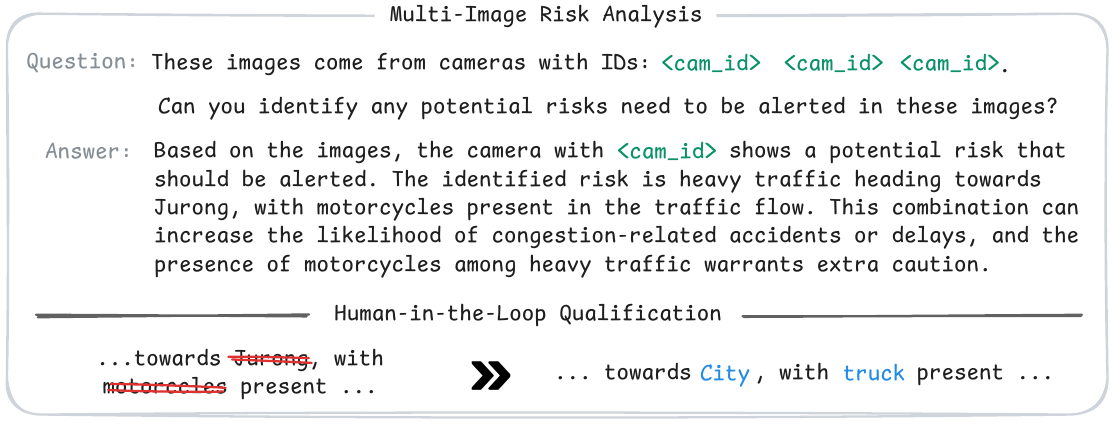}
    \caption{An instance of question-answer of multi-image risks analysis and corresponding human-in-the-loop amendment.}
    \label{pic:qa}
\end{figure*}

% ------------- LTA Evaluation Table ---------
\setlength{\tabcolsep}{3pt}
\begin{table*}[t]

\caption{Summary of existing vision--language datasets in intelligent transportation systems. The upper section presents datasets for question answering (QA), the middle section covers grounding-related tasks such as video grounding and multi-object tracking, and the lower section summarizes unified datasets.
We introduce LTD, the first open-ended traffic-domain VQA dataset. HITL: Human-in-the-loop; Temp.: Template; MOT: Multi-object tracking; ST-OG: Spatio-Temporal object grounding; MOG: Multi-object grounding.
}

\centering
\small
\setlength{\tabcolsep}{2mm}
\resizebox{\textwidth}{!}{
\begin{tabular}{l | c c c c c c}
\toprule
Dataset & Types & Annotation & \# QAs/Captions & \# Grounding & Domain & Data Format \\
\midrule
DRAMA & Video QA & HITL & 102k & - & Driving & Template \\
LingoQA & Video QA & HITL & 419k & - & Driving & Open-ended \\
NuScenes-QA & Image QA & Temp. & 460k & - & Driving & Template \\
DriveLM & Image QA & Temp. + HITL & 4.2M & - & Driving & Graph-Structured \\
CODA-LM & Image QA & Temp. + VLM + HITL & 9.8k & - & Driving & Open-ended \\
Omnidrive & Image QA & Temp. + VLM + HITL & 288k & - & Driving & Open-ended \\
\midrule
Refer-KITTI & MOT & HITL & - & 818 & Driving & - \\
NuPrompt & MOT & VLM & - & 35k & Driving & - \\
\midrule
SUTD-TrafficQA & Video QA & HITL & 62.5K & - & Traffic & Template \\
TUMTraffic-VideoQA & Video QA, ST-OG & Temp. + VLM & 87.3k & 5.7k & Traffic & Template \\
LTD (Ours) & Image QA, MOG & VLM + HITL & 8.8k & 2.8k & {\cellcolor[gray]{.9} Traffic} & {\cellcolor[gray]{.9}Open-ended} \\
\bottomrule
\end{tabular}
}
% \end{adjustbox}
\label{tab:ltd}
\end{table*}

\section{Land Transportation Dataset and Training Strategy}
\label{sec3}

\subsection{Data Annotation and Curation}\label{curation}

The universal overview of land transportation dataset (LTD) annotation and curation is illustrated in Fig. \ref{datacuration}. The images of the land transportation dataset are collected from multiple road cameras located at transportation network all over Singapore. As one can see from the Fig. \ref{datacuration}, the dataset consists of a broad spectrum of perspectives, for instance, diverse road structures including but not limited to highway, intersection, irregular junction, and various road users such as motorcycles, cyclists, pedestrians, vehicles, buses, trucks and so on. Additionally, the images are collected on a daily basis over day and night; therefore, a variety of weather and light conditions are all included. Then, this data collection is followed by the process of data filtering, where the blurry and blank scenes due to technical issues are filtered out in order to keep the high quality of the vision information. We first feed the filtered images together with the designed questions into pre-trained VLMs to generate task-specific responses. Recognizing that hallucinations are unavoidable in current open-source foundation models, we employ multiple VLMs to improve annotation robustness. For annotation, particularly in multi-image risk analysis, the two VLMs are prompted to produce open-ended reasoning responses, which are subsequently cross-validated to mitigate model-specific biases.
To further enhance annotation quality, we introduce a human-in-the-loop validation stage in which multiple examiners review the automatically generated annotations and refine counterfactual reasoning, as illustrated in Fig.~\ref{pic:qa}. For open-ended risk analysis tasks, we further apply a top-$k$ selection strategy over examiner responses to promote answer consistency and reliability.

After finalization, LTD contains 11.6K question--answer pairs in total, as summarized in Table~\ref{tab:ltd}.
As shown in the table, most existing vision--language datasets in intelligent transportation systems primarily focus on autonomous driving scenarios, whereas SUTD-TrafficQA, TUMTraffic-VideoQA, and LTD address city-scale traffic environments. Although the number of QA pairs in SUTD-TrafficQA and TUMTraffic-VideoQA is larger than that in LTD, the majority of their annotations adopt multiple-choice formats or are constructed from predefined templates. In contrast, LTD emphasizes open-ended reasoning, encouraging models to acquire traffic knowledge without relying on elimination strategies or dataset-specific biases. In particular, the multi-image risk analysis task further distinguishes LTD from prior datasets.
The visual inputs are captured by cameras located at different road segments, requiring VLMs to jointly reason over multiple images that may share minimal or no inherent correlation. Moreover, the risk analysis task requires not only identifying risky objects and contributing factors through open-ended reasoning, but also explicitly determining the risky road direction depicted in the images, which increases the overall reasoning complexity.

\subsection{Land Transportation Dataset Statistics}\label{sec:statistic}

\begin{figure*}[t]
    \centering
    \includegraphics[width=1.\linewidth]{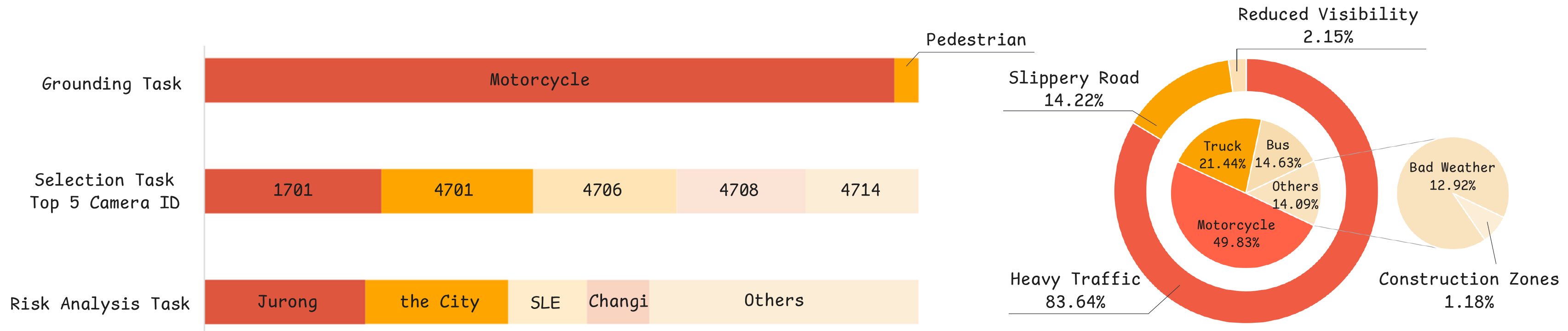}
    \caption{Statistical distributions of land transportation dataset.}
    \label{pic:statistics}
\end{figure*}

LTD comprises three categories of question--answer pairs: fine-grained multi-object grounding, multi-image camera ID selection, and open-ended multi-image risk analysis. The first task is single-image based, whereas the latter two involve three images per question. Figure~\ref{pic:statistics} summarizes the statistical distribution of LTD across these tasks. For the multi-object grounding task, we restrict the target categories to motorcycles and pedestrians, corresponding to critical vulnerable road users in traffic scenarios.
The number of motorcycle instances is substantially higher than that of pedestrians, reflecting the relatively low frequency of pedestrian appearances in traffic surveillance scenes. The term ``multi-object'' denotes that multiple instances may appear simultaneously within a single scene, including different combinations of motorcycles and pedestrians.

For the camera ID selection task, LTD focuses on five critical road sections: Moulmein Flyover LP448F (camera 1701), Alexander Road Exit (camera 4701), NUS School of Computing TID (camera 4706), ITE College West Dover TID (camera 4708), and West Coast Walk (camera 4714).
These locations include complex traffic structures such as deformed intersections, highway ramp merging and splitting areas, and construction zones. For the risk analysis task, the most frequently involved road directions are toward Jurong, the City, Seletar Expressway, and Changi.
The pie chart in Fig.~\ref{pic:statistics} further illustrates the distribution of multi-image risk analysis data in terms of risky factors (outer ring) and risky objects (inner ring). In addition to common traffic risks such as heavy traffic conditions (83.98\%) involving multiple road users, LTD also includes potential hazards associated with environmental factors, including slippery roads (13.46\%) and reduced visibility (2.15\%) caused by heavy rainfall.

\subsection{Architecture}\label{model}
The UniVLT architecture follows the standard design of the Qwen series, consisting of a redesigned Vision Transformer (ViT) and a decoder-only large language model (LLM), with minimal modifications to accommodate our task setting.

\paragraph{Vision Encoder.} In order to handle different native resolution inputs from various domain datasets, we employ a redesigned ViT as the vision encoder of UniVLT. From the input side, we unify all the resolution with a maximum of 225792 pixels, dynamically resizing the images larger than this resolution without destoration while smaller images are processed without padding, perserving their original size. We keep 2D Rotary Positional Embedding (RoPE) for position encoding and adopt RMSNorm for normalization. 

\paragraph{LLM.} The LLM backbone of UniVLT is initialized with pre-trained weights from LLM part of the Qwen2.5-VL. We keep the original settings including multimodal 1D RoPE (MRoPE), Grouped Query Attention (GQA)~\citep{gqa}, as well as RMSNorm~\citep{rmsnorm}.

For multi-image inputs, visual tokens from all images are concatenated in temporal or camera order, enabling the model to establish cross-image relationships and reason over long-range visual dependencies. Therefore, the visual input $X_v$ are organized as:
\begin{equation}
\begin{aligned}
    X_v = \{&\texttt{<img>}, v^1_1, \dots, v^1_N, \texttt{</img>}, \\ 
    &\texttt{<img>}, v^2_1, \dots, v^2_M, \texttt{</img>},\\
     &\qquad \qquad \ \ \ \dots \\
    &\texttt{<img>}, v^i_1, \dots, v^i_K, \texttt{</img>}\},
\end{aligned}
\end{equation}

\noindent where $v$ indicates visual tokens, $<img>$ and $</img>$ are special tokens for images, $i$ depicts the i-th image, N, M, K are the number of visual tokens of the three input images. 

Correspondingly, given a sequence of length $L$ for an answer text, the instruction prompts $X_q$ and target answers $X_a$ are denoted as:
\begin{equation}
\begin{aligned}
    X_q &= \{\texttt{<im\_start>}, \ell_1, \dots, \ell_Q, \texttt{</im\_end>} \}\\
    X_a &= \{\texttt{<im\_start>}, \ell_1, \dots, \ell_L, \texttt{</im\_end>} \},
\end{aligned}
\end{equation}

\noindent where $\ell$ depicts text tokens, $<im\_start>$ and $</im\_end>$ are special tokens for language, and Q indicates the number of textural tokens of the instruction prompt. Eventually, we compute the probability of the target answer by:
\begin{equation}
    p(X_a|X_v, X_q) = \prod_{l=1}^{L}p(x_i|X_v, X_{q, <i}, X_{a, <i}),
\end{equation}

\noindent where $X_{q, <i}$ and $X_{a, <i}$ are the instruction and answer tokens in all turns before the current prediction token $x_i$, respectively. 

\begin{figure*}[t]
    \centering
    \includegraphics[width=0.9\linewidth]{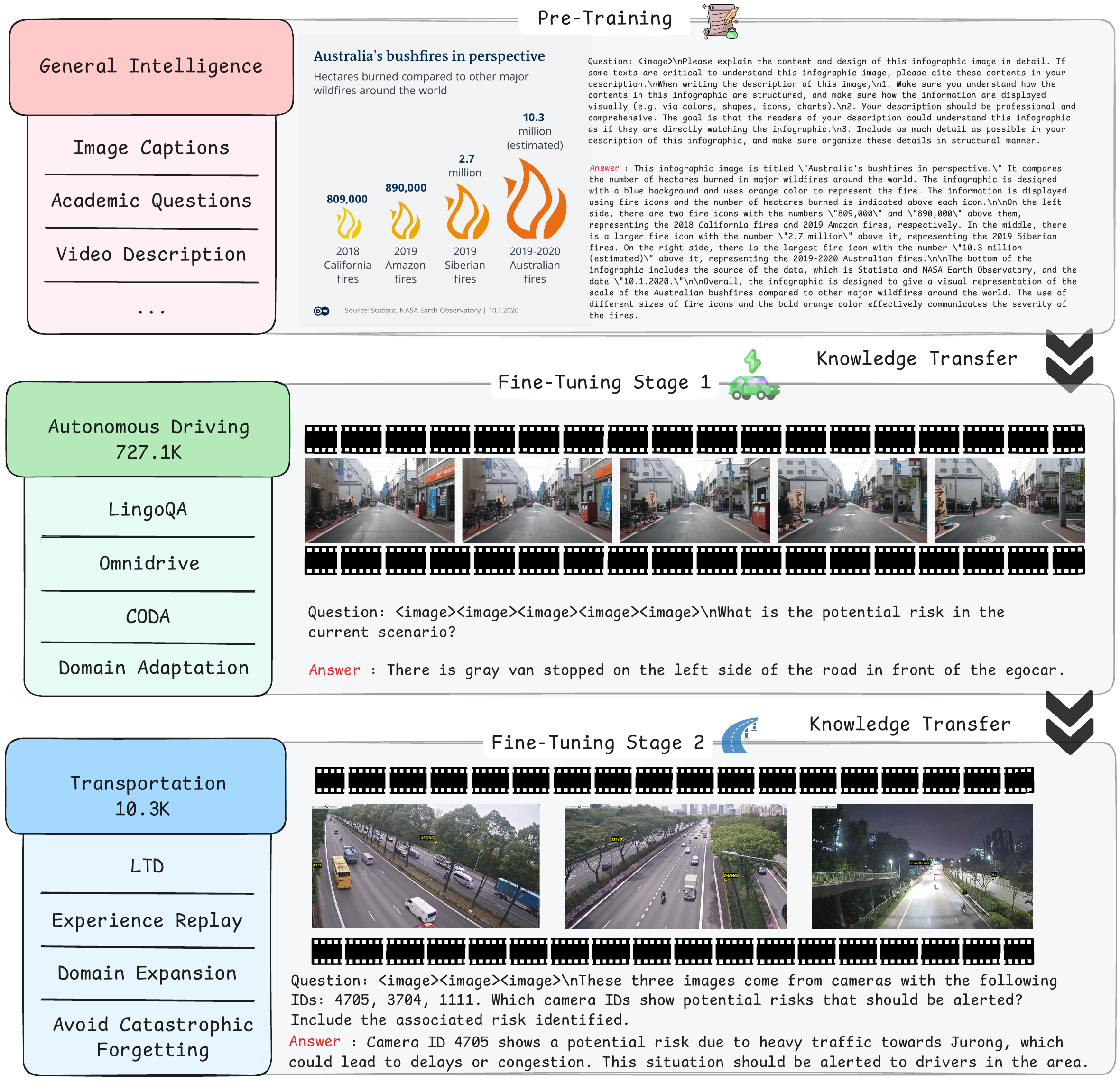}
    \caption{Training strategy and implementation details.
We adopt a curriculum-based knowledge transfer paradigm that progressively transitions from general-domain intelligence to autonomous driving, and subsequently to the traffic domain. To mitigate catastrophic forgetting of autonomous driving knowledge during the final adaptation stage, we incorporate an experience replay mechanism.
}
    \label{pic:recipe}
\end{figure*}

\subsection{Training Recipe}

In this section, we present the training recipe and strategy of the proposed UniVLT. The knowledge learning of UniVLT consists of three stages, including a pre-training stage followed by two fine-tuning phases. The rationale behind this design is that domain-specific knowledge can be substantially enhanced by leveraging general common sense, as they are strongly interconnected. Therefore, we adopt a multi-stage fine-tuning approach with LoRA~\citep{lora} technique to progressively transfer and adapt knowledge, leading to improved model performance across stages.

\noindent \textbf{Pre-Training Stage.} In this stage, UniVLT does not undergo additional training but instead inherits general common knowledge, such as textual reasoning, mathematical computation, and visual understanding, from large-scale internet datasets through the initialization of the well pre-trained Qwen2.5-VL 7B model~\citep{qwen2.5-vl}. We adopt Qwen2.5-VL as the backbone since it supports multi-image visual understanding, a crucial capability for the multi-image risk analysis tasks in UniVLT.

\noindent \textbf{Fine-Tuning Stage 1.} This stage focuses on transferring knowledge to the AD domain. The goal of this stage is to align the model’s general reasoning capability with domain-specific perception and decision-making knowledge. In this study, we employ three widely used open-source datasets: LingoQA~\citep{lingoqa}, OmniDrive~\citep{omnidrive}, and CODA~\citep{coda}. These datasets cover diverse tasks such as object detection, object grounding, scene understanding, driving suggestion, risk analysis, and action justification, resulting in a total of 727.1K QA pairs associated with either single-frame or multi-frame visual inputs.

\noindent \textbf{Fine-Tuning Stage 2.} In this stage, we expand the domain knowledge from autonomous vehicles (AVs) to the macroscopic traffic domain, which emphasizes holistic traffic analysis rather than ego-centric understanding. We utilize the 11.6K LTD dataset for the domain adaptation process and additionally include a subset of 3,000 samples from each AV dataset to mitigate catastrophic forgetting during fine-tuning.

A concrete example of each training stage is illustrated in Fig.~\ref{pic:recipe}. As shown, in the general VQA stage, each question is paired with a single image. In contrast, during Fine-Tuning Stage 1, each question is associated with five consecutive frames, providing temporal context for understanding motion and scene evolution. For the LTD stage, however, three images are sampled from three different cameras, meaning they are independent and share no spatial or temporal relationships. Consequently, the visual complexity increases progressively from the general VQA to the LTD stage. It is worth noting that multi-image risk analysis is particularly challenging, as it requires the VLM to establish meaningful connections between multiple images and accurately extract key information relevant to the question.

\section{Experiment}
\label{sec4}

% ------------- LTA Evaluation Table ---------
\setlength{\tabcolsep}{3pt}
\begin{table*}[tp]

\caption{Quantitative results of UniVLT compared with other open-source pre-trained and AD-tailored VLMs on the LTD benchmark. $^{\dagger}$\emph{Grounding evaluation protocol.} Since different models adopt model-specific bounding box output formats, we report results only for those that produce normalized coordinates or whose outputs can be converted to a thousandth-level normalized scale.}
\begin{adjustbox}{width=1.0\textwidth,center}

\centering
\small
\setlength{\tabcolsep}{2mm}
\begin{tabular}{l l | c | c | c}
\toprule
\multirow{2}{*}{Model} & \multirow{2}{*}{Size} &
\multicolumn{1}{c|}{Multi-Image Risk Analysis} &
\multicolumn{1}{c|}{Camera ID Selection} &
\multicolumn{1}{c}{Multi-Object Grounding} \\
& & GPT-Score$\uparrow$ & Accuracy$\uparrow$ & F1 Score$^{\dagger}$$\uparrow$ \\
\midrule
LLaVA-OV & 0.5B & 0.01 & 0.29 & 0.00 \\
LLaVA-OV & 7B   & 0.03 & 0.32 & 0.00 \\
Qwen2.5-VL & 7B & 0.46 & 0.48 & 0.45 \\
InternVL2.5 & 8B & 0.25 & 0.23 & 0.00 \\
Qwen3-VL & 4B & 0.14 & 0.25 & 0.62 \\
\midrule
OpenEMMA & 3B & 0.10 & 0.29 & 0.44 \\
WiseAD & 1.7B & 0.00 & 0.00 & N/A \\
RoboTron-Drive & 8B & 0.06 & 0.00 & 0.06 \\
ReCogDrive & 8B & 0.29 & 0.32 & N/A \\
\midrule
UniVLT & 7B & \textbf{0.66} & \textbf{0.66} & \textbf{0.64} \\
\bottomrule
\end{tabular}
\end{adjustbox}
\label{tab:lta}
\end{table*}

\subsection{Dataset}
In this work, we employ three open-source autonomous driving datasets, namely LingoQA~\citep{lingoqa}, CODA-LM~\citep{coda}, and OmniDrive~\citep{omnidrive}, together with a traffic dataset introduced in this work, referred to as LTD. LingoQA, released by Wayve, comprises 413,800 samples covering a wide range of tasks, including object recognition, scene description, driving-related reasoning, and action justification. CODA-LM extends visual question answering toward corner-case scenarios and contains 25,400 samples. OmniDrive further enriches driving knowledge related to traffic rules and counterfactual reasoning, providing 287,900 samples. Finally, LTD is a multi-image dataset consisting of 10,300 samples that capture road segments and traffic scenes across Singapore.

\subsection{Baselines}
To thoroughly evaluate the superiority of the proposed UniVLT, we employ multiple baselines, including both general and AD-tailored VLMs. The general VLMs include LLaVA-OV~\citep{llava-ov}, Qwen2.5-VL~\citep{qwen2.5-vl}, Qwen3-VL~\citep{qwen3vl}, and InternVL2.5~\citep{internvl2.5}, while the AD-specific category comprises state-of-the-art (SOTA) approaches such as OpenEMMA~\citep{openemma}, WiseAD~\citep{wisead}, RoboTron-Drive~\citep{drivemm}, and ReCogDrive~\citep{recogdrive}. It is worth noting that we are unable to compare with transportation-tailored VLMs, as no open-source models are currently available. 

\subsection{Metrics}
We utilize standard NLP metrics, including CIDEr~\citep{cider}, METEOR~\citep{meteor}, and BLEU~\citep{bleu}, to comprehensively evaluate reasoning quality and alignment with human judgment. For grounding tasks, we employ the F1 score instead of language-based metrics, since the results are based on coordinate predictions with a fixed answer pattern. For datasets that provide their own evaluation metrics, we additionally report the corresponding results, such as LingoJudge for LingoQA~\citep{lingoqa} and GPT-based evaluation for CODA-LM~\citep{coda} and OmniDrive~\citep{omnidrive}.

\subsection{Results on Land Transportation Dataset}\label{sec:results_ltd}

% ------------- LingoQA Evaluation Table ---------
\begin{figure*}[t]
    \centering
    \includegraphics[width=.95\linewidth]{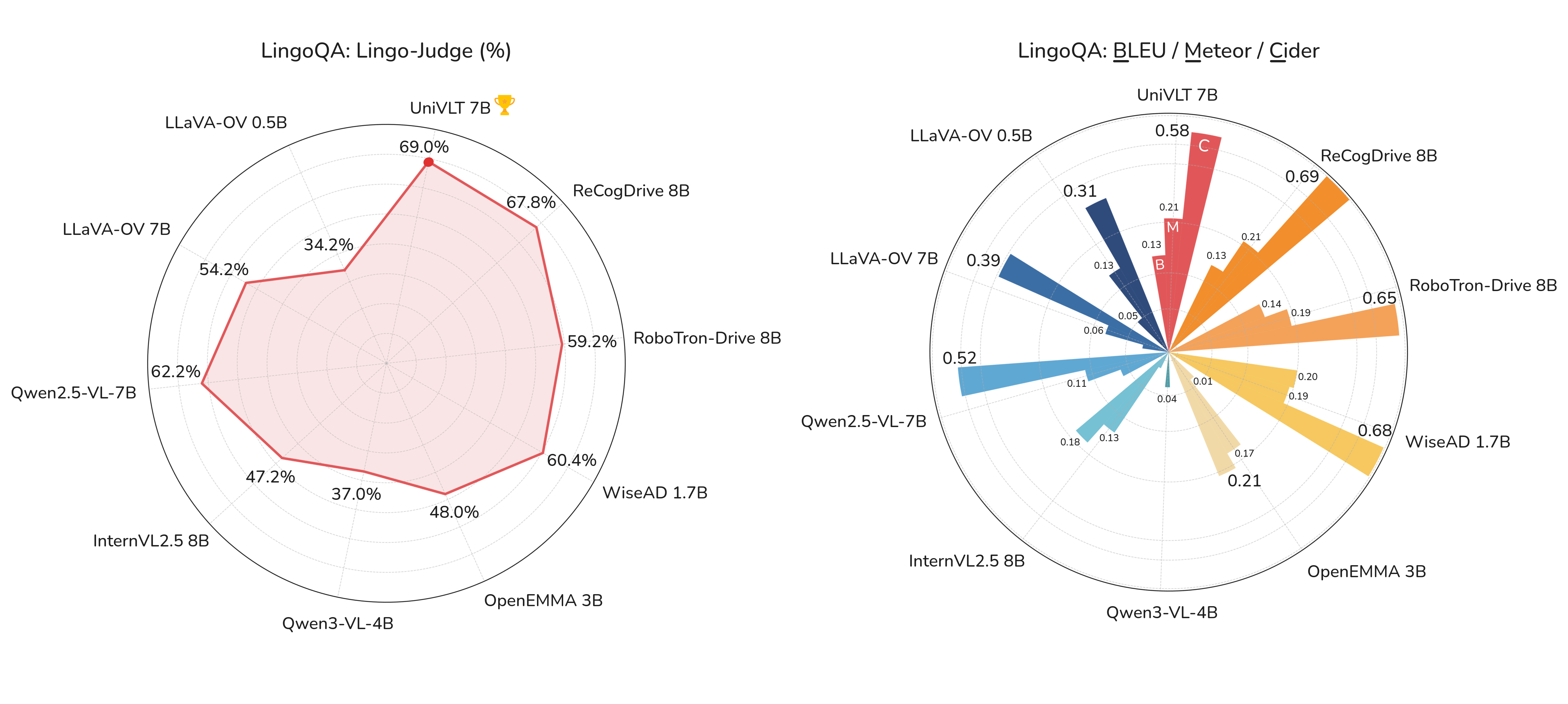}
    \caption{Quantitative result of UniVLT compared with other open-source pre-trained and AD-tailored VLMs on LingoQA benchmark.}
    \label{pic:lingoqa}
\end{figure*}

We first evaluate UniVLT over the proposed LTD which primarily focuses on three tasks, \textit{multi-image risk analysis}, \textit{camera id selection}, and \textit{multi-object grounding}. Quantitative results are reported in Table~\ref{tab:lta}. 

UniVLT consistently outperforms all baseline models by a substantial margin across all evaluation settings. On the \textit{multi-image risk analysis task}, UniVLT achieves a GPT score of 0.66, whereas SOTA pre-trained VLMs, including Qwen2.5-VL and Qwen3-VL, obtain scores of at most 0.46 and 0.14, respectively. This significant performance gap indicates that UniVLT possesses a markedly stronger capability for comprehensive scene understanding and risk-aware reasoning. It is worth noting that \textit{multi-image risk analysis} is inherently challenging, as it requires not only recognizing and reasoning about potential hazards in complex driving scenarios, but also accurately identifying road directions, risky objects, and the underlying contributing factors.

For the \textit{camera id selection} task, UniVLT achieves an accuracy of 0.66, again outperforming all competing methods by a substantial margin.
Interestingly, we observe that general-purpose pre-trained VLMs tend to outperform AD-tailored VLMs on this task.
We attribute this performance gap to the characteristics of our dataset, in which multi-image inputs are collected from different road cameras and therefore exhibit no explicit temporal or spatial correlations across images.
This setting contrasts sharply with the strongly correlated multi-frame or surrounding-view data that are commonly used to fine-tune AD-tailored VLMs.

For the \textit{multi-object grounding} task, we report results only for models that produce normalized coordinates or whose outputs can be converted to a thousandth-level normalized scale, since different models adopt model-specific bounding box output formats. As shown in Table~\ref{tab:lta}, UniVLT achieves an F1 score of 0.64, outperforming AD-tailored VLMs, the backbone model Qwen2.5-VL, and the recently released pre-trained Qwen3-VL (0.62). This outcome is expected, as grounding is a general vision task that largely depends on the quality of backbone pre-training, whereas the other two tasks are more domain-specific. Notably, several strong baselines achieve an F1 score of 0.0, which can be attributed to the characteristics of infrastructure-mounted camera imagery, where vulnerable road users often occupy very small regions, making grounding highly sensitive to minor localization errors. Qualitative results of UniVLT on LTD are provided in Appendix~\ref{fig:lta-vis} and Appendix~\ref{fig:lta-vis-moto}.

Overall, these results demonstrate that UniVLT effectively handles not only the fine-grained perception task, but also deals with multi-image risk analysis under realistic traffic conditions. The large performance margin of UniVLT against other pre-trained and AD-tailored VLMs further highlights the significance of the LTD from the vision (multiple non-correlated images) and traffic domain perspective, and confirms the feasibility and necessity of UniVLT for the practical application. As for the effectiveness of the training recipe, we conduct a detailed ablation studies and provide the results and corresponding analysis in Section~\ref{ablation}.

\begin{figure*}[t]
    \centering
    \includegraphics[width=.95\linewidth]{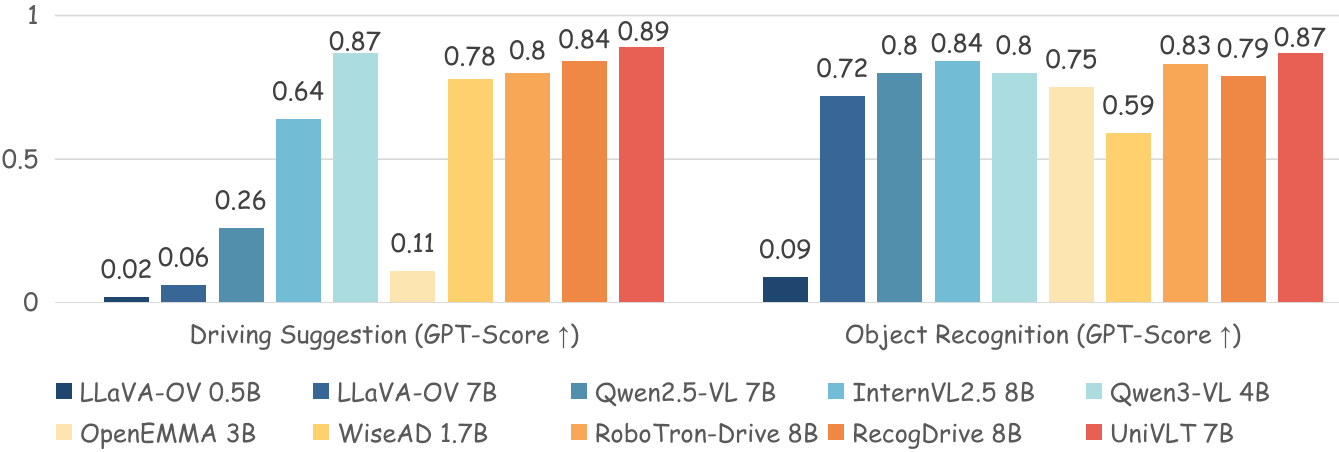}
    \caption{Quantitative result of UniVLT compared with other open-source pre-trained and AD-tailored VLMs on OmniDrive dataset.}
    \label{pic:omnidrive}
\end{figure*}

\subsection{Results on LingoQA Dataset}\label{sec:results_lingoqa}
We further evaluate the proposed UniVLT on the LingoQA~\citep{lingoqa} dataset, which comprises diverse AD-related tasks, including object recognition, scene understanding, driving suggestion, and action justification.
Quantitative results are reported in Fig.~\ref{pic:lingoqa}. As shown in the table, UniVLT consistently outperforms all baseline methods and achieves a SOTA of 69.0\% on the official benchmark metric, Lingo-Judge. Specifically, UniVLT exceeds the pre-trained model LLaVA-OV 7B~\citep{llava-ov} by 14.8\% and the AD-tailored VLM ReCogDrive~\citep{recogdrive} by 1.2\%, demonstrating its strong multitasking and reasoning capabilities even within the AD domain.

It is worth noting that all pre-trained VLMs, as well as OpenEMMA~\citep{openemma}, are evaluated in a zero-shot setting, whereas the remaining AD-tailored VLMs report in-domain performance after being fine-tuned on LingoQA.
Interestingly, despite zero-shot evaluation, Qwen2.5-VL 7B~\citep{qwen2.5-vl} attains a competitive score of 62.2\%, highlighting its strong reasoning ability stemming from general-purpose pre-training. In addition to the Lingo-Judge metric, we also report standard NLP metrics for all methods to provide complementary insights from a textual evaluation perspective. Qualitative results of UniVLT on LingoQA dataset are provided in Appendix~\ref{fig:lingqa-vis}.

Overall, the SOTA results on the LingoQA dataset validate the essential multitasking and reasoning capabilities of the proposed UniVLT. 

\subsection{Results on OmniDrive Dataset}\label{sec:results_omnidrive1}
We further evaluate our method on the planning-oriented driving knowledge dataset OmniDrive~\citep{omnidrive}, which includes traffic rule answering and counterfactual reasoning tasks. Quantitative results are summarized in Fig.~\ref{pic:omnidrive}.
Since OmniDrive does not provide an official evaluation metric, we adopt the GPT score as the primary performance indicator, together with standard NLP metrics.

For the \textit{object recognition} task, several pre-trained VLMs achieve competitive GPT scores despite relatively low performance on standard NLP metrics.
This discrepancy highlights the limitation of token-level metrics in capturing semantic equivalence, as distinct textual descriptions may convey the same underlying meaning.
In contrast, GPT-based evaluation better reflects semantic alignment, under which models such as InternVL2.5 and the Qwen-VL series attain comparable performance.
UniVLT achieves the best results under both evaluation protocols, indicating robust scene understanding from both token-level and semantic-level perspectives.

For \textit{planning-oriented driving suggestion}, the performance gap between pre-trained and AD-tailored models becomes more pronounced. The GPT-score results show that most AD-tailored VLMs substantially outperform pre-trained, underscoring the importance of incorporating domain-specific driving knowledge beyond general intelligence. While Qwen3-VL 4B~\citep{qwen3vl} achieves the highest GPT score among pre-trained VLMs, UniVLT attains the best overall performance and consistently outperforms all baselines across all evaluation metrics.
Qualitative results of UniVLT on Omnidrive are provided in Appendix~\ref{fig:omnidrive-obj-recog-vis} and Appendix~\ref{fig:omnidrive-driving-sugg-vis}. 

\subsection{Results on CODA Dataset}\label{sec:results_coda}
Last but not least, the UniVLT is also evaluated under the three tasks of CODA~\citep{coda} dataset focusing on challenging corner-case scenarios. We follow the official benchmark metric, namely CODA-LM, and report the quantitative result in Table~\ref{tab:coda}.

As shown in the table, UniVLT achieves the best performance on the \emph{General Perception} task with a CODA-LM score of 5.18, while attaining the second-highest score on \emph{Region Perception} (7.25) and competitive performance on \emph{Driving Suggestion} (5.62). These results demonstrate that UniVLT effectively balances open reasoning of high-level scene understanding and low-level action planning in complex and safety-critical scenarios.

We note that RoboTron-Drive attains a higher score on the \textit{Region Perception} task. This performance gap can be partially attributed to its use of additional data augmentation on CODA dataset from 37K to 184K during training~\citep{drivemm}. In contrast, UniVLT is trained without curating extra augmented data, in order to remain faithful to the original data distribution and to avoid overfitting to specific scenes or corner cases. 

For the \textit{Driving Suggestion} task, the pre-trained SOTA model, namely Qwen3-VL, achieves the highest CODA-LM score among all baselines. This result suggests that driving suggestion generation may benefit from strong general-purpose pre-training, as the task closely relates to planning-oriented reasoning that integrates perceived scene context into action recommendations. However, Qwen3-VL exhibits relatively lower performance on both region-level and general perception tasks compared to UniVLT and RoboTron-Drive, indicating that strong driving suggestion performance does not necessarily require the most accurate fine-grained perception. Qualitative results of UniVLT on CODA-LM are provided in Appendix~\ref{fig:codalm-vis-region-percep}, Appendix~\ref{fig:codalm-vis-general-percep} and Appendix~\ref{fig:codalm-vis-driving-suggestion}.

Overall, UniVLT demonstrates strong and well-balanced performance across all three CODA tasks. Despite being trained without additional data augmentation, UniVLT achieves the best performance on the \textit{General Perception} task and remains highly competitive on \textit{Region Perception} and \textit{Driving Suggestion}. 

%------------------CODA-----------------------
% ------------- CODA Evaluation Table (CODA-LM only) ---------
\begin{table*}[tp]
\caption{Quantitative result of UniVLT compared with other open-source pre-trained and AD-tailored VLMs on CODA-LM benchmark. Best results are in \textbf{bold}, second best are \underline{underlined}.}
\begin{adjustbox}{width=1.0\textwidth,center}
\centering
\small
\setlength{\tabcolsep}{3mm}
\begin{tabular}{l l | c c c c}
\toprule
\multicolumn{2}{c|}{} &
\multicolumn{3}{c}{\textbf{CODA-LM}$\uparrow$} \\
\cmidrule(l){3-6}
Model & Size &
\textbf{Region Perception} &
\textbf{General Perception} &
\textbf{Driving Suggestion} & 
\textbf{Average}\\
\midrule
LLaVA-OV & 0.5B & 1.93 & 1.11 & 1.57 & 1.54 \\
LLaVA-OV & 7B & 3.06 & 1.33 & 2.68 & 2.36 \\
Qwen2.5-VL & 7B & 5.07 & 4.37 & 5.20 & 4.88 \\
InternVL2.5 & 8B & 6.04 & 3.91 & 5.22 & 5.06 \\
Qwen3-VL & 4B & 6.92 & 4.54 & \textbf{6.08} & 5.85 \\
\midrule
OpenEMMA & 3B & 5.52 & 3.76 & 5.04 & 4.77 \\
WiseAD & 1.7B & 1.81 & 1.27 & 1.09 & 1.39 \\
RoboTron-Drive & 8B & \textbf{7.66} & \underline{5.15} & \underline{5.68} & \textbf{6.16} \\
ReCogDrive & 8B & 6.94 & 5.08 & 5.58 & 5.87 \\
\midrule
UniVLT & 7B & \underline{7.25} & \textbf{5.18} & 5.62 & \underline{6.02} \\
\bottomrule
\end{tabular}
\end{adjustbox}
\label{tab:coda}
\end{table*}

\section{Ablation Studies}\label{ablation}
We conduct systematic ablation studies to examine the impact of the key design choices in UniVLT, with a particular focus on the proposed curriculum-style training recipe and domain adaptation strategy. Specifically, we compare our training settings against (i) direct fine-tuning with LTD solely, (ii) direct joint-training with AD and LTD, and (iii) multi-stage fine-tuning without AD experience replay. Unless otherwise stated, all ablation models share the same backbone, optimization settings, and evaluation protocols.

\subsection{Impact of Curriculum Training Strategy}
We first evaluate different training strategies on the LTD benchmark. Specifically, we consider two direct training baselines, where one model is fine-tuned solely on LTD and the other is jointly fine-tuned on both AD and LTD. In contrast, our approach adopts a multi-stage training strategy, in which the model is first adapted to the AD domain in the first stage and then transferred to LTD in the second stage with a limited amount of AD experience replay. The quantitative results are summarized in Table~\ref{tab:ablation_ltd}.

We observe that the baseline model directly fine-tuned on LTD achieves comparable performance on camera id identification and multi-object grounding, but performs poorly on the multi-image risk analysis task. Joint training on both AD and LTD improves the reasoning quality for multi-image risk analysis; however, it significantly degrades performance on the other two tasks. This performance drop is likely caused by naively mixing heterogeneous domains, which introduces conflicting inductive biases.

In contrast, our approach adopts a two-stage curriculum transfer training strategy and achieves substantial performance gains over the direct training baselines, yielding the best overall performance across all tasks. These results confirm both the necessity and effectiveness of the proposed curriculum transfer training scheme, which progressively transfers knowledge from the general domain to the application-specific domain, thereby enhancing scene understanding and open-ended reasoning capabilities.

% --------------- Ablation LTA ------------------------
\begin{table*}[t]
\renewcommand\arraystretch{1.5}
\caption{Ablation results on the LTD test set.
We analyze the impact of curriculum transfer and domain adaptation (AD).} 
\begin{adjustbox}{width=1.0\textwidth,center}
\centering
\small
\setlength{\tabcolsep}{4.8pt}
\begin{tabular}{c c cc cc ccc c}
\toprule
\multirow{2}{*}{Ablation Attribute} &
\multirow{2}{*}{Curriculum Training} &
\multicolumn{2}{c}{Fine-Tuning Stage 1} &
\multicolumn{2}{c}{Fine-Tuning Stage 2} &
\multicolumn{3}{c}{LTD} &
\multirow{2}{*}{LingoQA} \\
\cmidrule(lr){3-4}\cmidrule(lr){5-6}\cmidrule(lr){7-9}
& & AD data & LTD data & AD data & LTD data & Risk Analysis & Camera ID & Grounding & \\
\midrule
 Directly Tailored for LTD & \xmark & \xmark & \checkmark & \xmark & \xmark & 0.44 & 0.63 & 0.62 & N/A\\
 Joint Training & \xmark  & \checkmark & \checkmark & \xmark & \xmark & 0.60 & 0.53 & 0.49 & 67.6 \\
 Catastrophic Forgetting & \checkmark & \checkmark & \xmark & \xmark & \checkmark & 0.65 & 0.65 & \textbf{0.65} & 59.8\\
 Ours & \checkmark & \checkmark & \xmark & \checkmark & \checkmark & \textbf{0.66} & \textbf{0.66} & 0.64 & \textbf{69.0}\\
\bottomrule
\end{tabular}
\end{adjustbox}
\label{tab:ablation_ltd}
\end{table*}

\subsection{Impact of Experience Replay}
To evaluate the impact of experience replay in the second stage of our training strategy, we further compare our method with a baseline that follows the same two-stage fine-tuning procedure but excludes AD data during the second-stage adaptation. This ablation study aims to quantitatively assess the contribution of experience replay in mitigating catastrophic forgetting.

As shown in the last two rows of Table~\ref{tab:ablation_ltd}, although the baseline achieves comparable performance on the LTD benchmark, its reasoning performance on the LingoQA~\citep{lingoqa} dataset degrades significantly. Specifically, the model suffers from severe catastrophic forgetting, with an almost 10\% drop in Lingo-Judge metrics. These results demonstrate the effectiveness of the experience replay strategy employed in our framework, as it plays a critical role in stabilizing knowledge transfer across training stages.

\section{Conclusion}
\label{sec6}
In this work, we introduce the Land Transportation Dataset (LTD), a city-scale vision--language dataset designed to facilitate open-ended reasoning in urban traffic environments. By integrating fine-grained multi-object grounding, multi-image camera identification, and multi-image open-ended risk analysis, LTD provides a comprehensive benchmark for evaluating macroscopic traffic-level reasoning. The dataset emphasizes minimally correlated multi-view inputs and open-ended annotations, encouraging models to move beyond shortcut-based inference towards structured reasoning over complex traffic scenarios. Building upon LTD, we further propose UniVLT, a transportation foundation model trained via curriculum-based knowledge transfer. UniVLT unifies knowledge across autonomous driving and traffic domains within a single architecture, while preserving stability and scalability. Extensive experiments demonstrate that UniVLT achieves strong performance across diverse open-ended reasoning tasks and reveals limitations of existing foundation models under multi-image and cross-view settings. As future work, we will explore broader cross-city generalization and more diverse practical application-oriented tasks. We expect LTD and UniVLT to serve as practical benchmarks for advancing multimodal open-ended reasoning in intelligent transportation systems.

\clearpage

\bibliographystyle{unsrtnat}
\bibliography{references}

\clearpage
\onecolumn
\appendix

\section{Additional Visualizations}
\label{app:viz}
\FloatBarrier

\setcounter{figure}{0}
\renewcommand{\thefigure}{S\arabic{figure}}

\begin{figure}[ht]
  \centering

  % --- Images ---
  \begin{subfigure}[t]{0.32\linewidth}
    \includegraphics[width=\linewidth]{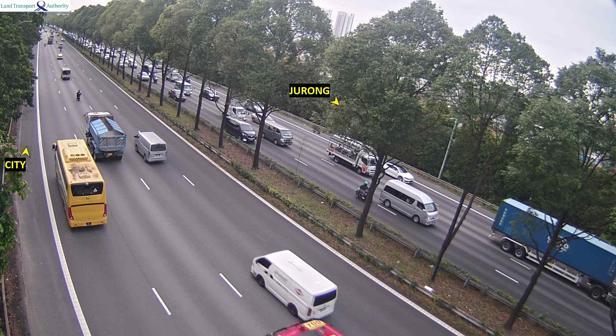}
    \caption{Camera 4705}
  \end{subfigure}
  \hfill
  \begin{subfigure}[t]{0.32\linewidth}
    \includegraphics[width=\linewidth]{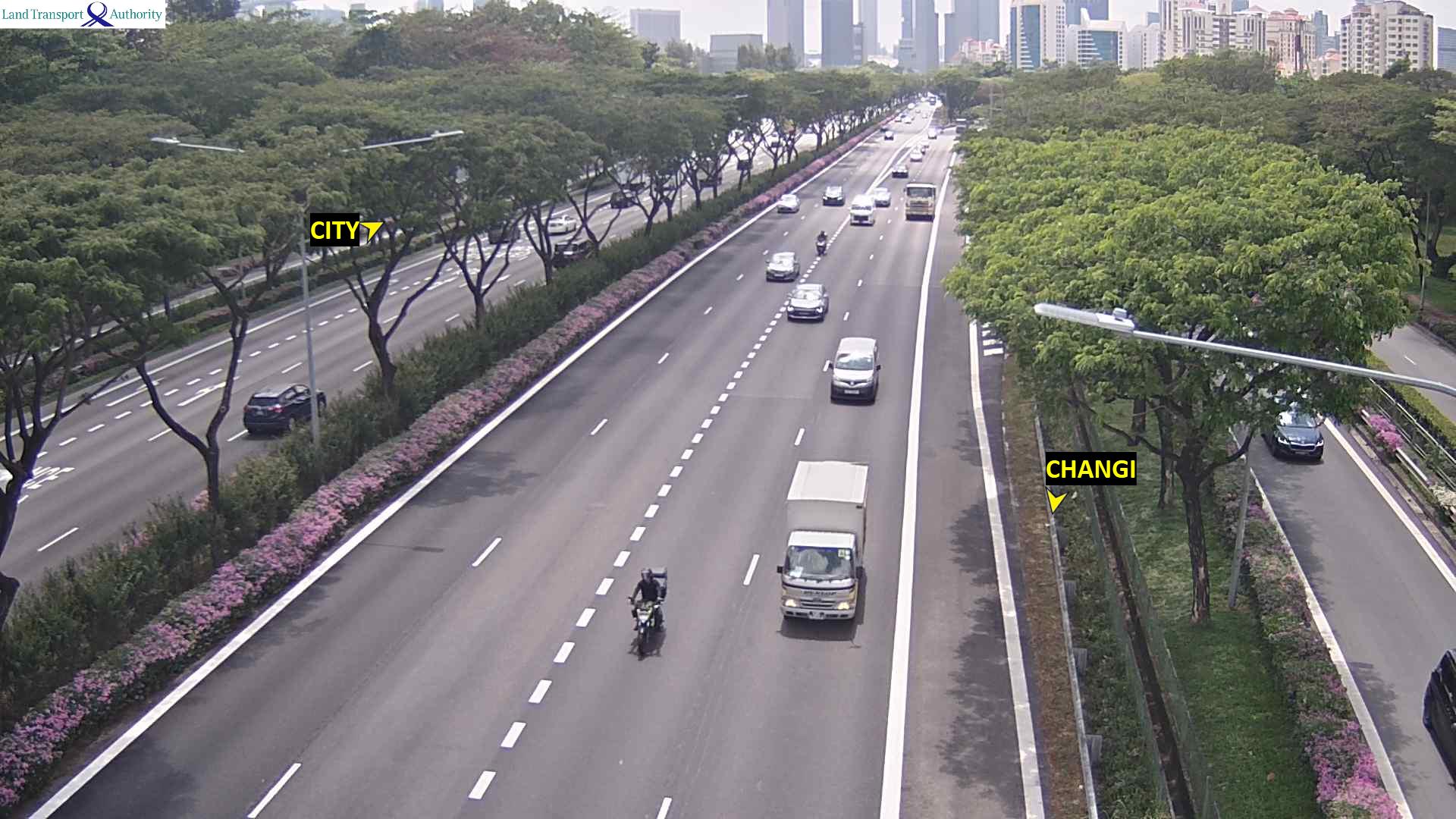}
    \caption{Camera 3704}
  \end{subfigure}
  \hfill
  \begin{subfigure}[t]{0.32\linewidth}
    \includegraphics[width=\linewidth]{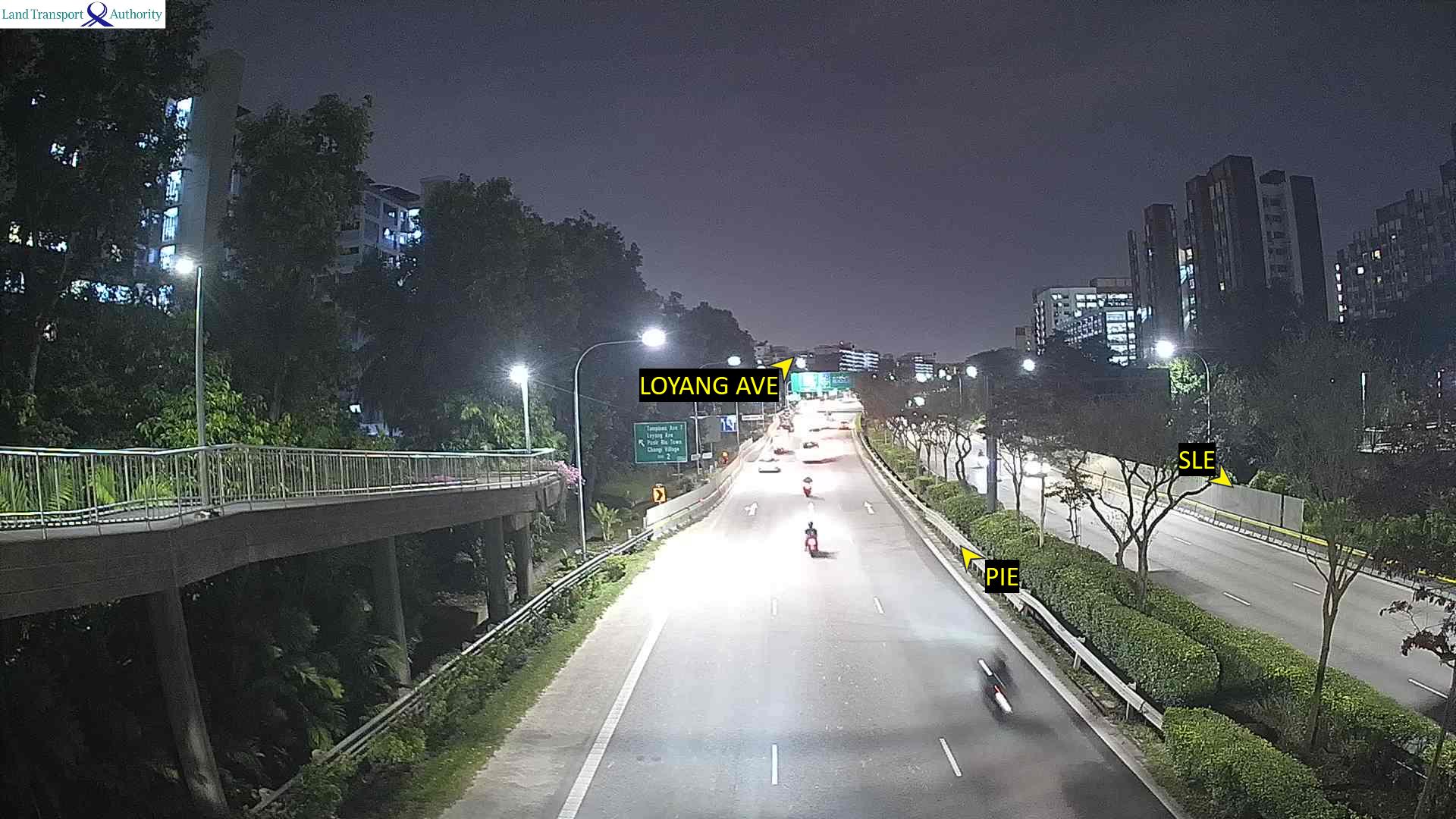}
    \caption{Camera 1111}
  \end{subfigure}

  \vspace{0.75em}

  % --- Table ---
  \renewcommand{\arraystretch}{1.2}
  \begin{tabularx}{0.95\linewidth}{@{}lX@{}}
    \toprule
    \textbf{User} & These three images come from cameras with the following IDs: 4705, 3704, 1111. Which camera IDs show potential risks that should be alerted? Include the associated risk identified. \\
    \midrule
    \textbf{UniVLT} & \textcolor{ForestGreen}{Camera ID 4705} shows a potential risk due to heavy traffic \textcolor{ForestGreen}{towards Jurong}, which could \textcolor{ForestGreen}{lead to delays or congestion}. This situation should be alerted to drivers in the area. \\
    \midrule
    \textbf{Ground Truth} & Based on the analysis, \textcolor{ForestGreen}{camera 4705} should be alerted due to the presence of heavy traffic heading \textcolor{ForestGreen}{towards Jurong}. The traffic includes motorcycles, buses, and trucks, increasing the \textcolor{ForestGreen}{risk of congestion and potential accidents}.\\
    \bottomrule
  \end{tabularx}

  \vspace{0.75em}

  \caption{Qualitative instance of LTD risk analysis.}
  \label{fig:lta-vis}
\end{figure}

\setcounter{figure}{\value{figure}}

% lta grounding visualization
\begin{figure}[ht]
  \centering

  % --- Image ---
  \begin{subfigure}[t]{0.45\linewidth}
    \includegraphics[width=\linewidth]{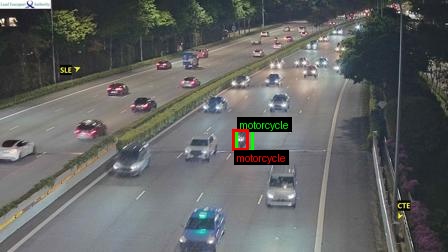}
    \caption{Camera 1111}
  \end{subfigure}

  \vspace{0.75em}

  % --- Table ---
  \renewcommand{\arraystretch}{1.2}
  \begin{tabularx}{0.95\linewidth}{@{}lX@{}}
    \toprule
    \textbf{User} & In this image, give coordinates [x\_min, y\_min, x\_max, y\_max] for the detected motorcycle. \\
    \midrule
    \textbf{UniVLT} & There is a motorcycle [234, 131, 246, 148]. \\
    \midrule
    \textbf{Ground Truth} & There is a motorcycle [237, 131, 251, 149]. \\
    \bottomrule
  \end{tabularx}

  \vspace{0.75em}

  \caption{Qualitative instance of motorcycle localization with predicted (\textcolor{red}{red}) and ground-truth (\textcolor{ForestGreen}{green}) bounding boxes.}
  \label{fig:lta-vis-moto}
\end{figure}

%------ LingoQA figure -----
\setcounter{figure}{\value{figure}}

\begin{figure}[ht]
  \centering

  % --- Images ---
  \begin{subfigure}[t]{0.32\linewidth}
    \includegraphics[width=\linewidth]{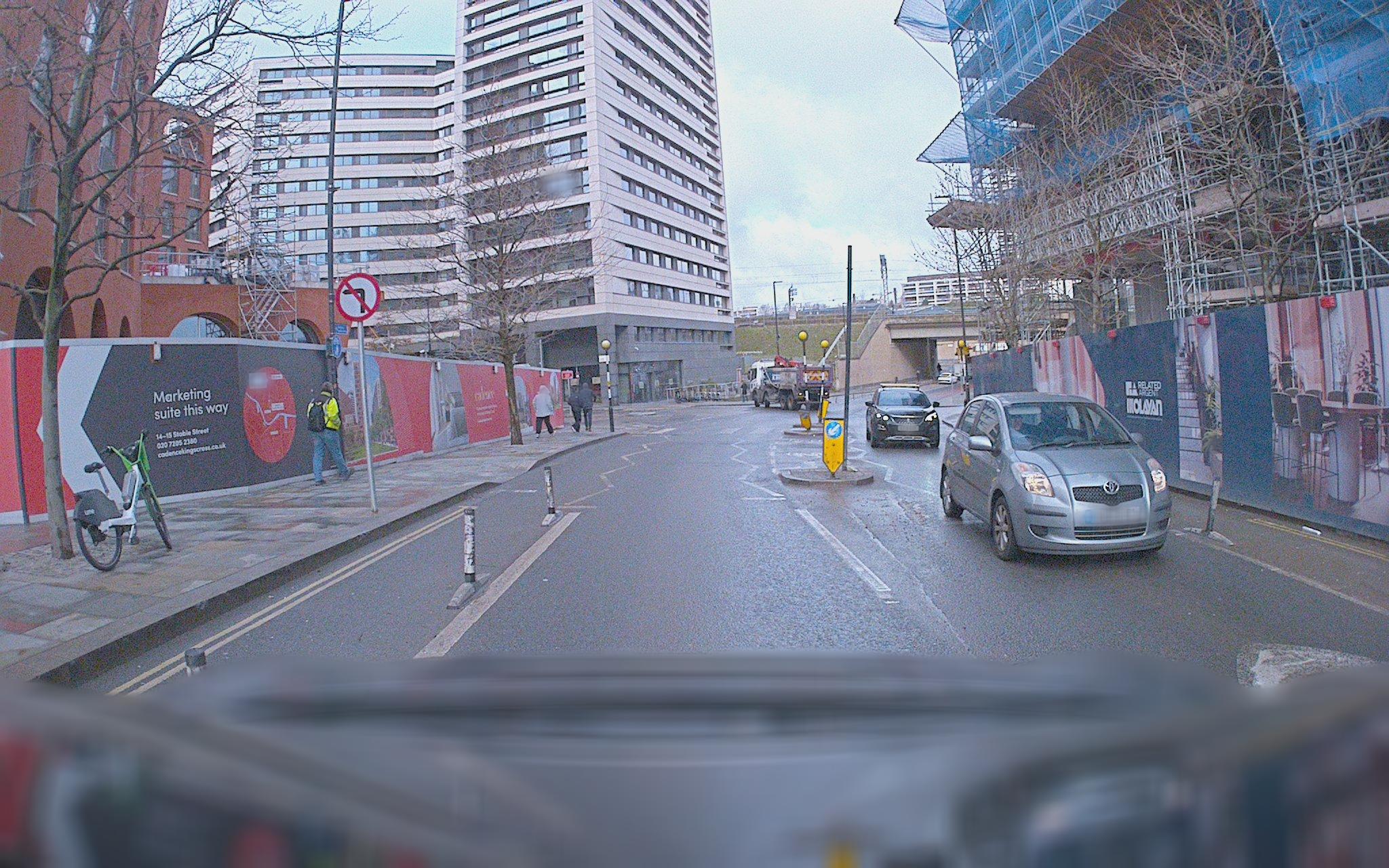}
    \caption{Frame 0}
  \end{subfigure}
  % \hfill
  \begin{subfigure}[t]{0.32\linewidth}
    \includegraphics[width=\linewidth]{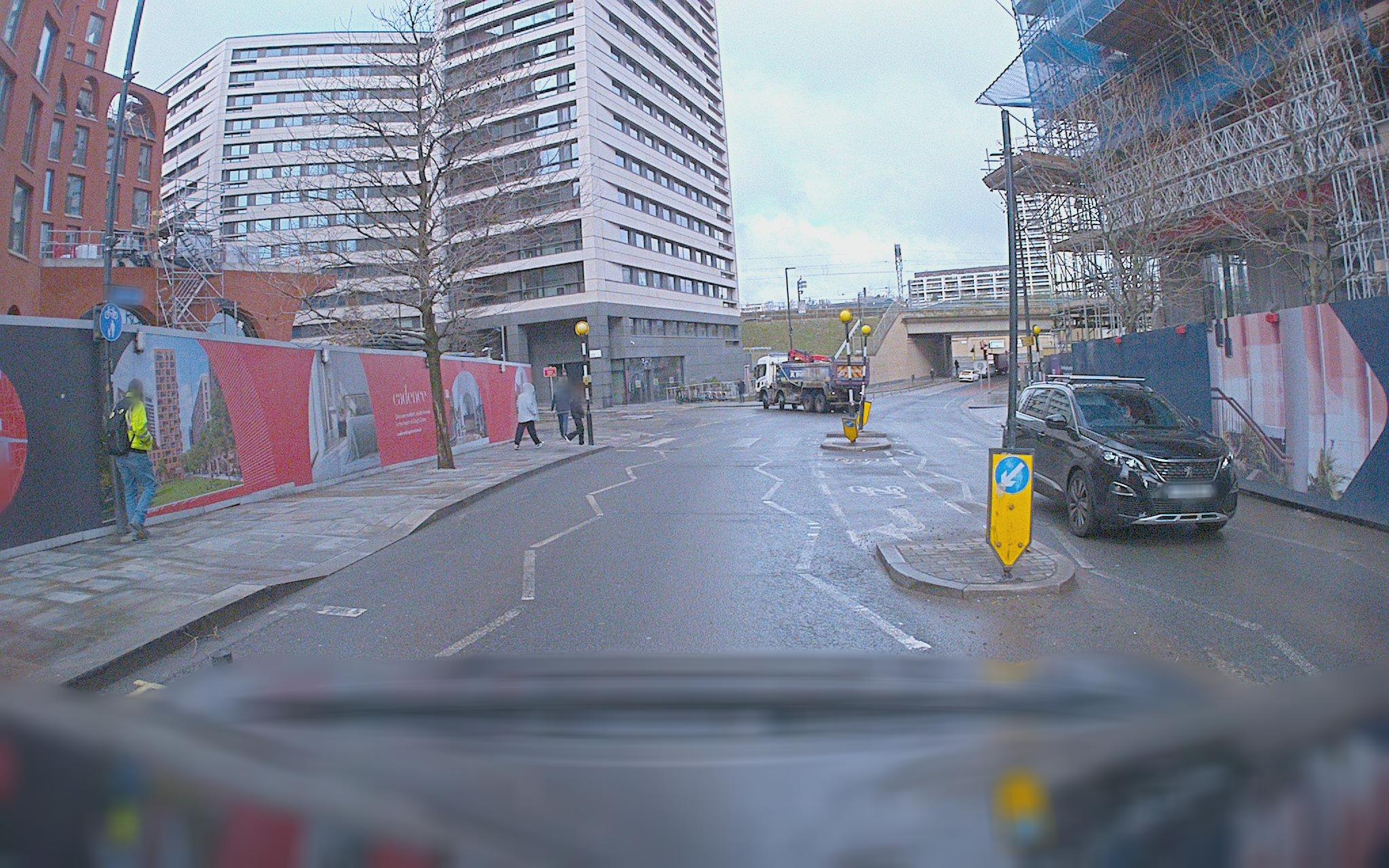}
    \caption{Frame 1}
  \end{subfigure}
  % \hfill
  \begin{subfigure}[t]{0.32\linewidth}
    \includegraphics[width=\linewidth]{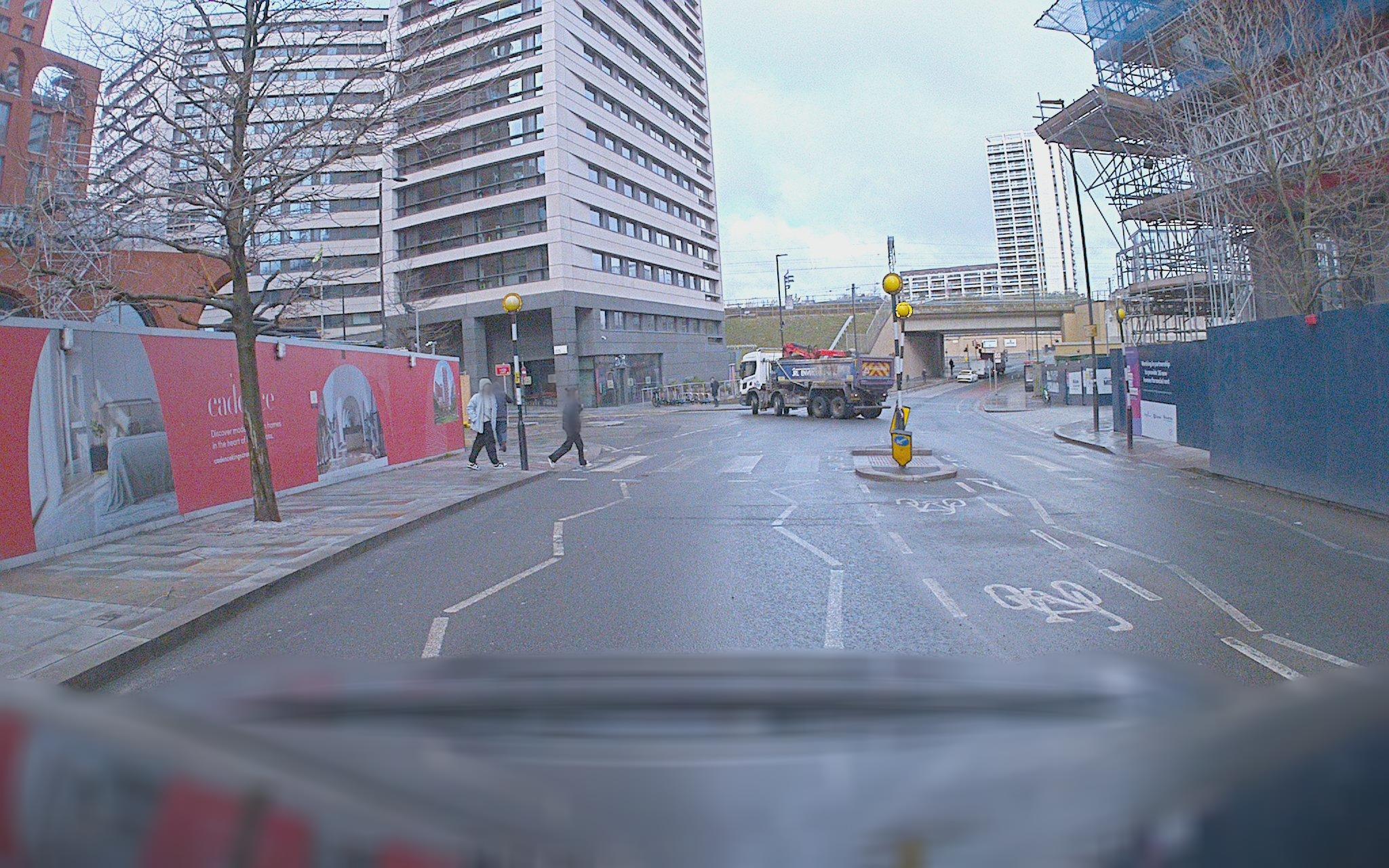}
    \caption{Frame 2}
  \end{subfigure}
  % \hfill
  \begin{subfigure}[t]{0.32\linewidth}
    \includegraphics[width=\linewidth]{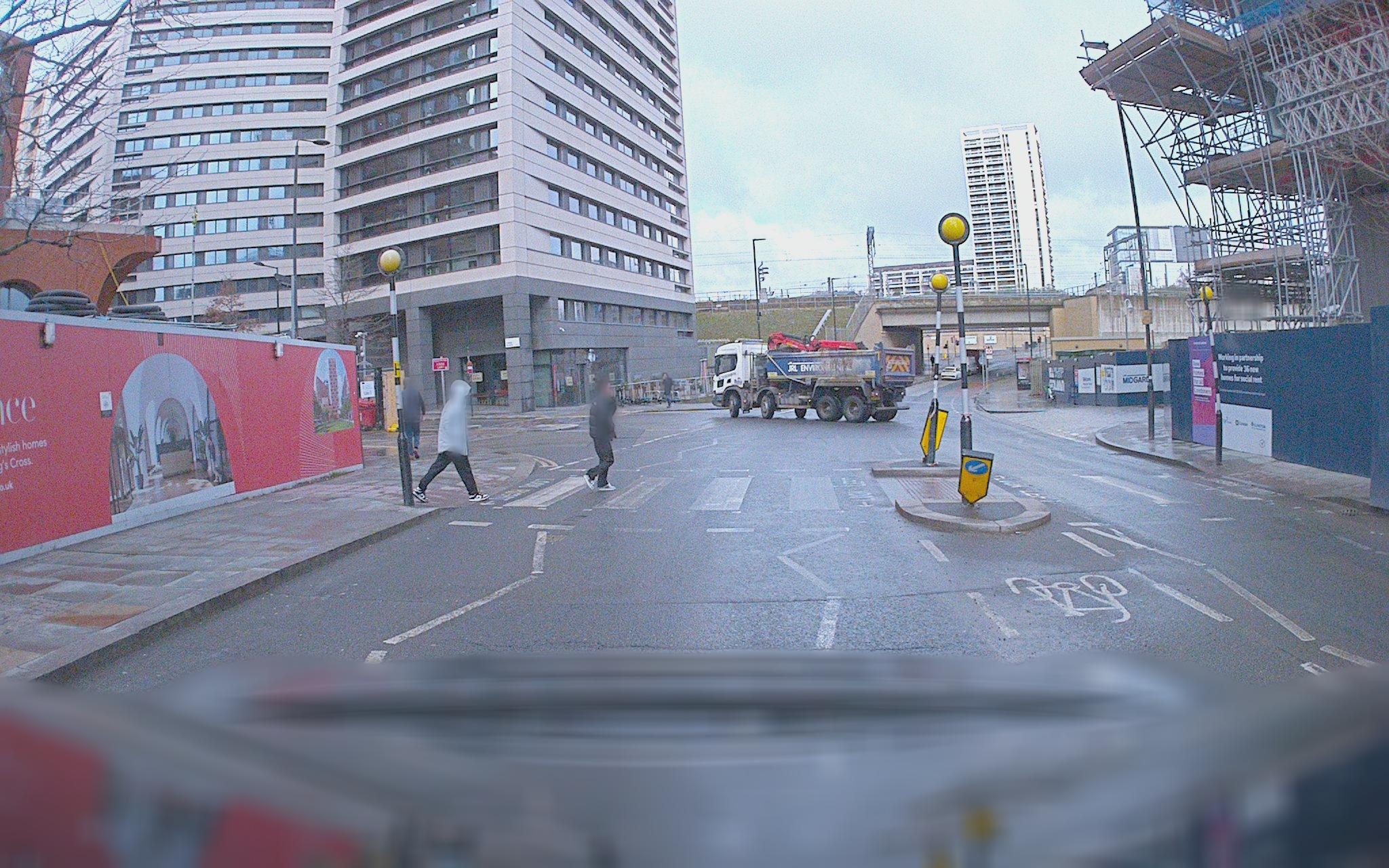}
    \caption{Frame 3}
  \end{subfigure}
  % \hfill
  \begin{subfigure}[t]{0.32\linewidth}
    \includegraphics[width=\linewidth]{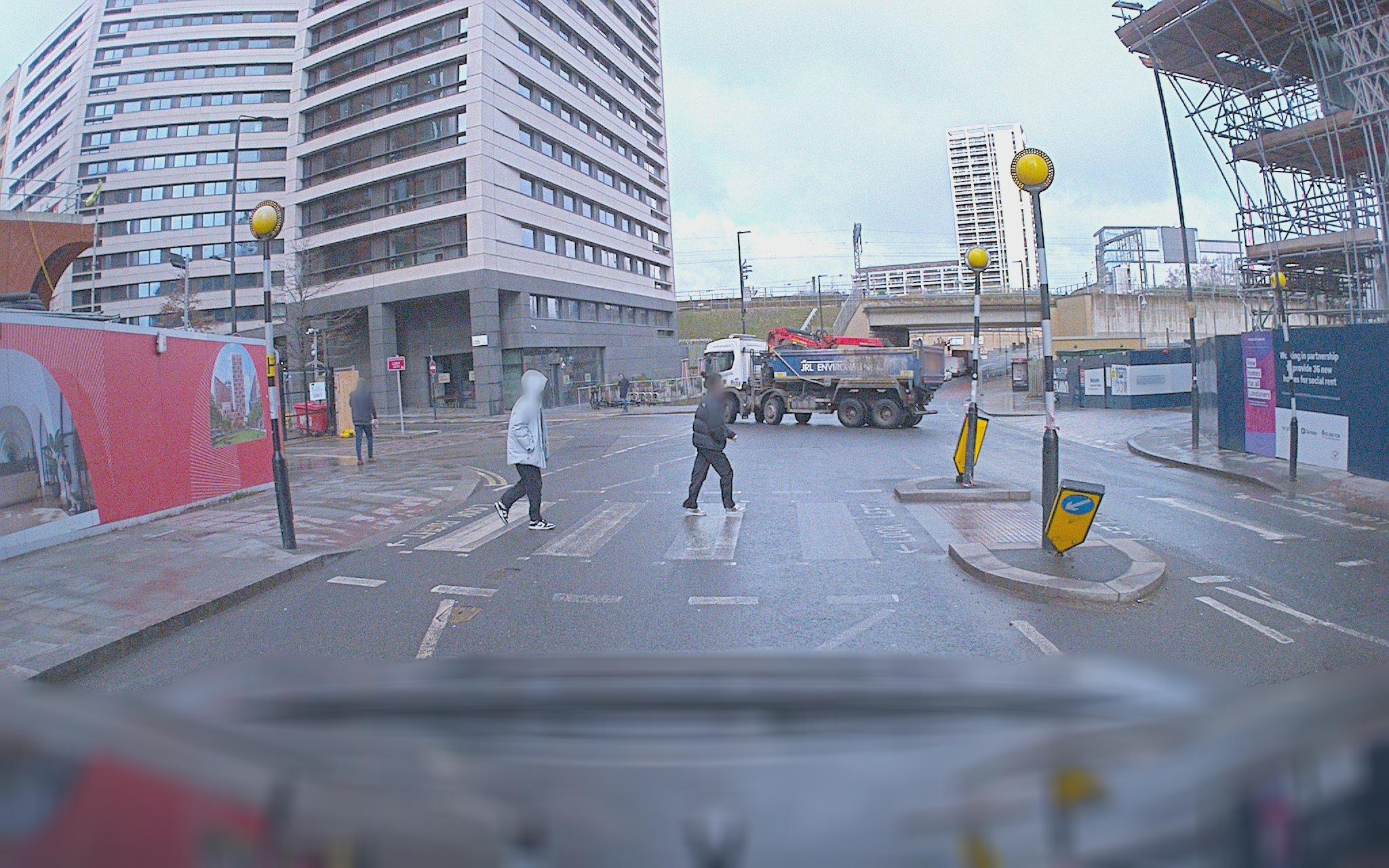}
    \caption{Frame 4}
  \end{subfigure}

  \vspace{0.75em}

  % --- Table ---
  \renewcommand{\arraystretch}{1.2}
  \begin{tabularx}{0.95\linewidth}{@{}lX@{}}
    \toprule
    \textbf{User} & What is your current driving action and it's justification? \\
    \midrule
    \textbf{UniVLT} & I am \textcolor{ForestGreen}{decelerating} to \textcolor{ForestGreen}{let the pedestrian safely cross} on the zebra or pedestrian crossing ahead. \\
    \midrule
    \textbf{Ground Truth} & \textcolor{ForestGreen}{Stopping}, to \textcolor{ForestGreen}{let pedestrians cross} at the zebra crossing. \\
    \bottomrule
  \end{tabularx}

  \vspace{0.75em}

  \caption{Qualitative instance of LingoQA.}
  \label{fig:lingqa-vis}
\end{figure}

% ----- CODA-LM region perception -----
\setcounter{figure}{\value{figure}}

\begin{figure}[t]
  \centering

  % --- Image ---
  \begin{subfigure}[t]{0.45\linewidth}
    \includegraphics[width=\linewidth]{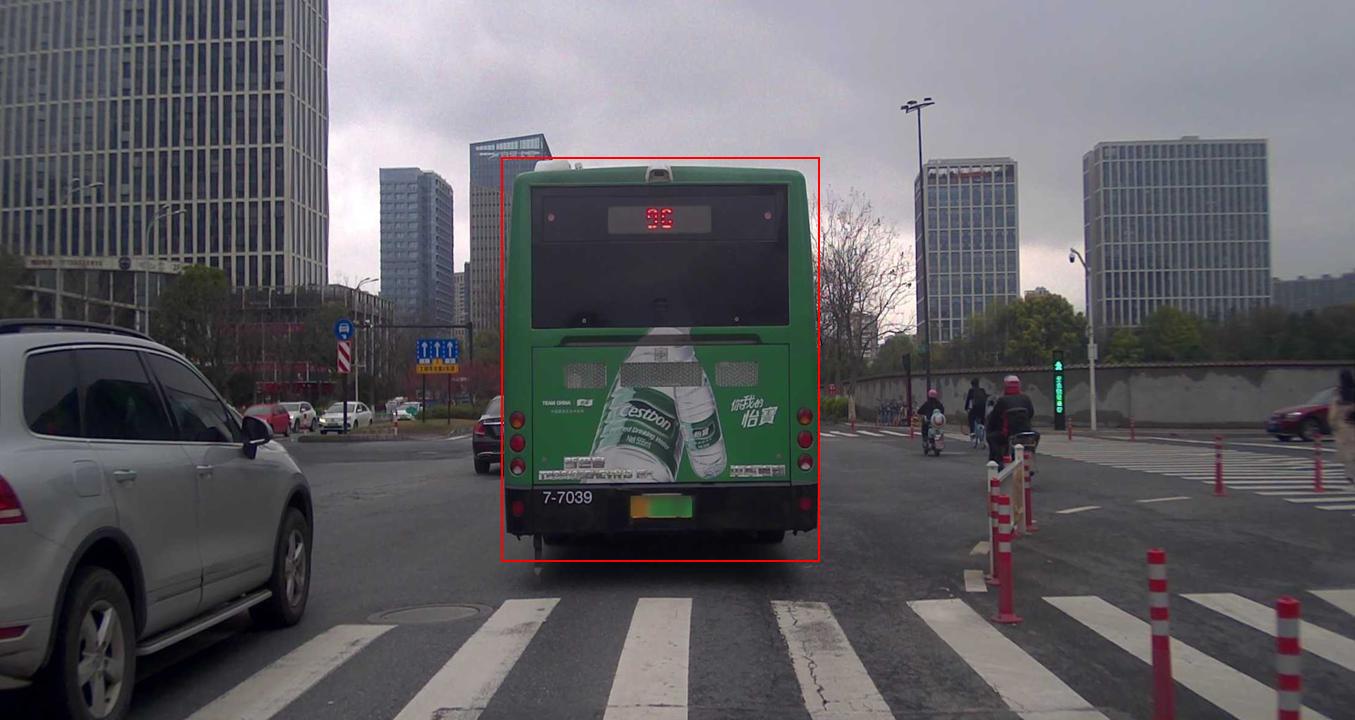}
    % \caption{Camera 1111}
  \end{subfigure}

  \vspace{0.75em}

  % --- Table ---
  \renewcommand{\arraystretch}{1.2}
  \begin{tabularx}{0.95\linewidth}{@{}lX@{}}
    \toprule
    \textbf{User} & Please describe the object inside the red rectangle in the image and explain why it affect ego car driving. \\
    \midrule
    \textbf{UniVLT} & This is a large green bus, identified by its size and design typical of public transport vehicles. The presence of this bus \textcolor{ForestGreen}{directly ahead} in the same lane as the ego car suggests that the ego car \textcolor{ForestGreen}{must maintain a safe following distance} and \textcolor{ForestGreen}{be prepared for potential stops or slow movement}, especially if the bus is \textcolor{ForestGreen}{approaching a bus stop} or navigating through traffic. \\
    \midrule
    \textbf{Ground Truth} & A large, motorized vehicle designed to carry multiple passengers. This vehicle's presence \textcolor{ForestGreen}{directly ahead} indicates that the ego car will \textcolor{ForestGreen}{need to maintain a safe following distance} and \textcolor{ForestGreen}{be prepared to stop or slow down} if the bus does, especially in a situation where it might \textcolor{ForestGreen}{pick up or drop off passengers}. \\
    \bottomrule
  \end{tabularx}

  \vspace{0.75em}

  \caption{Qualitative instance of CODA-LM Region Perception.}
  \label{fig:codalm-vis-region-percep}
\end{figure}

% ----- CODA-LM region perception -----
\setcounter{figure}{\value{figure}}

\begin{figure}[t]
  \centering

  % --- Image ---
  \begin{subfigure}[t]{0.45\linewidth}
    \includegraphics[width=\linewidth]{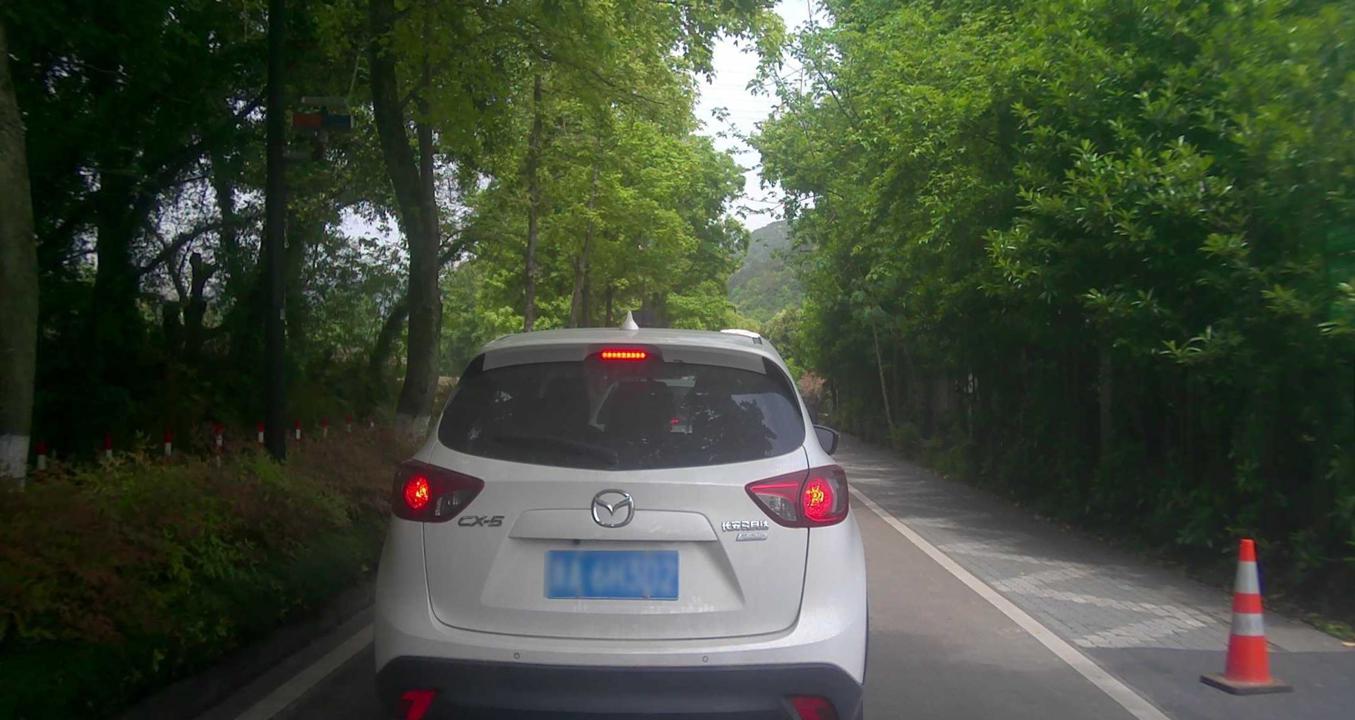}
    % \caption{Camera 1111}
  \end{subfigure}

  \vspace{0.75em}

  % --- Table ---
  \renewcommand{\arraystretch}{1.2}
  \begin{tabularx}{0.95\linewidth}{@{}lX@{}}
    \toprule
    \textbf{User} & There is an image of traffic captured from the perspective of the ego car. Focus on objects influencing the ego car's driving behavior: vehicles (cars, trucks, buses, etc.), vulnerable road users (pedestrians, cyclists, motorcyclists), traffic signs (no parking, warning, directional, etc.), traffic lights (red, green, yellow), traffic cones, barriers, miscellaneous(debris, dustbin, animals, etc.). You must not discuss any objects beyond the seven categories above. Please describe each object's appearance, position, direction, and explain why it affects the ego car's behavior. \\
    \midrule
    \textbf{UniVLT} & In the traffic scene observed, there is a \textcolor{red}{silver SUV} \textcolor{ForestGreen}{directly in front of the ego vehicle}, identifiable by its \textcolor{ForestGreen}{brake lights being illuminated}, which suggests that the vehicle is either \textcolor{ForestGreen}{slowing down or has come to a stop}. This situation necessitates that the ego car \textcolor{ForestGreen}{maintains a safe following distance and is prepared to halt if necessary}.
    Additionally, there is a \textcolor{red}{single orange traffic cone} positioned on the right side of the road. The presence of this cone \textcolor{ForestGreen}{indicates that there may be some form of road work or obstruction ahead}, prompting the ego car to \textcolor{ForestGreen}{exercise caution and possibly reduce speed} as it approaches this area. \\
    \midrule
    \textbf{Ground Truth} & In the traffic image under observation, there are two notable categories of road features: vehicles and traffic cones.
    Starting with the vehicle category, there is a \textcolor{red}{white SUV} located \textcolor{ForestGreen}{directly in front of our car}. The SUV's \textcolor{ForestGreen}{brake lights are illuminated}, signifying that the vehicle has either \textcolor{ForestGreen}{slowed down significantly or come to a complete halt}. This specific situation demands that our car \textcolor{ForestGreen}{maintains an appropriate following distance from the SUV, ensuring readiness to either stop or decelerate} further based on the future movements of the SUV. The careful consideration of the SUV's actions and positioning is crucial for ensuring safety and adhering to traffic regulations.
    Moving onto the traffic cones, there is a \textcolor{red}{line-up of orange and white traffic cones} positioned on the right side of the road. These cones are typically utilized to signify lane restrictions, which could be \textcolor{ForestGreen}{indicative of ongoing road construction} or the presence of obstacles outside the current lane of travel. The placement of these cones serves as a warning to drivers, emphasizing the need for heightened caution. Motorists are advised against entering the areas marked by these cones to avoid potential hazards and to maintain the orderly flow of traffic.
    In this traffic scenario, both the presence of the white SUV and the alignment of traffic cones have direct implications on driving behavior. They serve as indicators for drivers to exercise \textcolor{ForestGreen}{increased caution, adjust speeds, and prepare for possible changes} in the road's layout or traffic conditions. \\
    \bottomrule
  \end{tabularx}

  \vspace{0.75em}

  \caption{Qualitative instance of CODA-LM General Perception.}
  \label{fig:codalm-vis-general-percep}
\end{figure}

% ----- CODA-LM driving suggestion -----
\setcounter{figure}{\value{figure}}

\begin{figure}[t]
  \centering

  % --- Image ---
  \begin{subfigure}[t]{0.45\linewidth}
    \includegraphics[width=\linewidth]{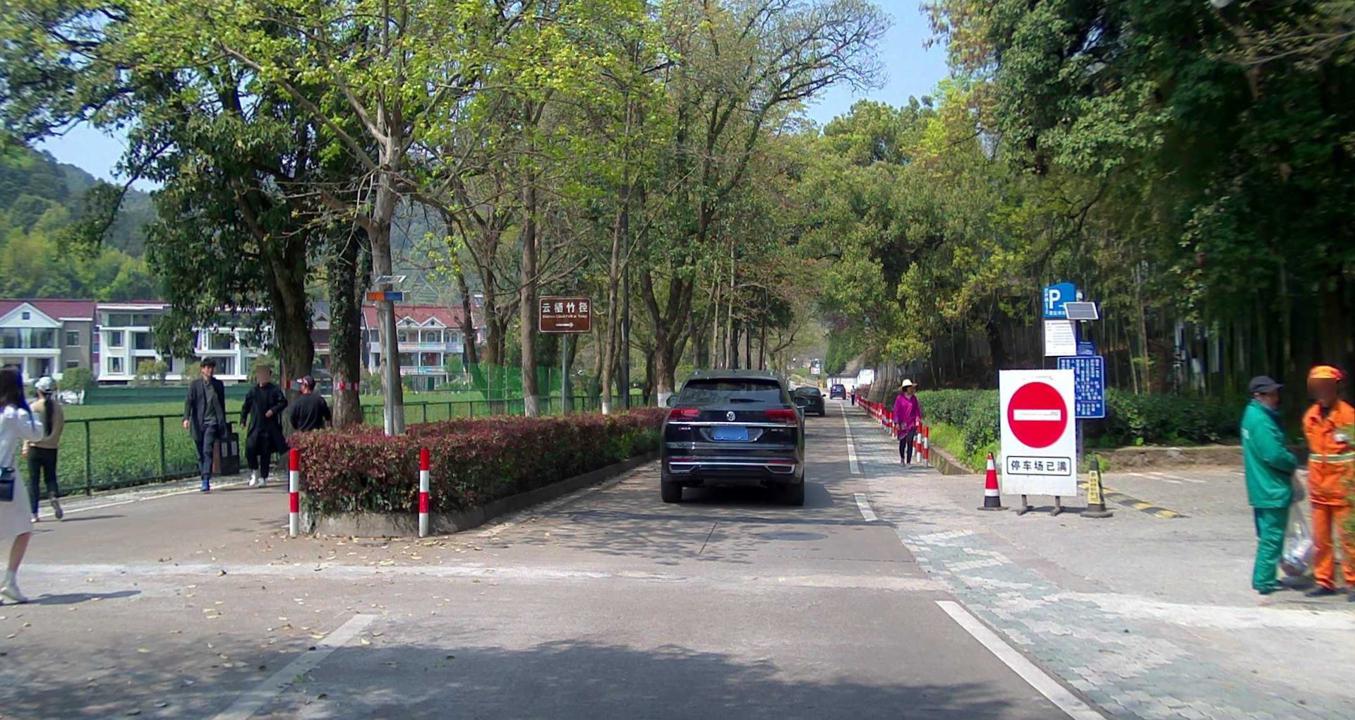}
    % \caption{Camera 1111}
  \end{subfigure}

  \vspace{0.75em}

  % --- Table ---
  \renewcommand{\arraystretch}{1.2}
  \begin{tabularx}{0.95\linewidth}{@{}lX@{}}
    \toprule
    \textbf{User} & There is an image of traffic captured from the perspective of the ego car. Focus on objects influencing the ego car's driving behavior: vehicles (cars, trucks, buses, etc.), vulnerable road users (pedestrians, cyclists, motorcyclists), traffic signs (no parking, warning, directional, etc.), traffic lights (red, green, yellow), traffic cones, barriers, miscellaneous(debris, dustbin, animals, etc.). You must not discuss any objects beyond the seven categories above. Please provide driving suggestions for the ego car based on the current scene. \\
    \midrule
    \textbf{UniVLT} & The ego car should \textcolor{ForestGreen}{maintain a safe following distance behind the black SUV ahead}, be prepared to stop or \textcolor{ForestGreen}{yield to pedestrians} crossing the street, and \textcolor{ForestGreen}{avoid entering the restricted area} marked by the \textcolor{ForestGreen}{'No Entry' sign}. Additionally, the driver should stay alert for any changes in the road conditions due to the traffic cones and be ready to navigate around them if necessary. \\
    \midrule
    \textbf{Ground Truth} & This vehicle should \textcolor{ForestGreen}{maintain a safe following distance from the SUV ahead}, always be ready to \textcolor{ForestGreen}{respond to pedestrian movements}, especially pay attention to the movements of workers on the right, and comply with \textcolor{ForestGreen}{no-entry traffic signs, not entering restricted lanes}. The vehicle should also monitor the road paths marked by traffic posts, and while maintaining a safe driving speed, be prepared to take necessary evasive actions to deal with any unexpected pedestrian or worker movements. \\
    \bottomrule
  \end{tabularx}

  \vspace{0.75em}

  \caption{Qualitative instance of CODA-LM Driving Suggestion.}
  \label{fig:codalm-vis-driving-suggestion}
\end{figure}

% ----- omnidrive general perception -----
\setcounter{figure}{\value{figure}}

\begin{figure}[t]
  \centering

  % --- Image ---
  \begin{subfigure}[t]{0.45\linewidth}
    \includegraphics[width=\linewidth]{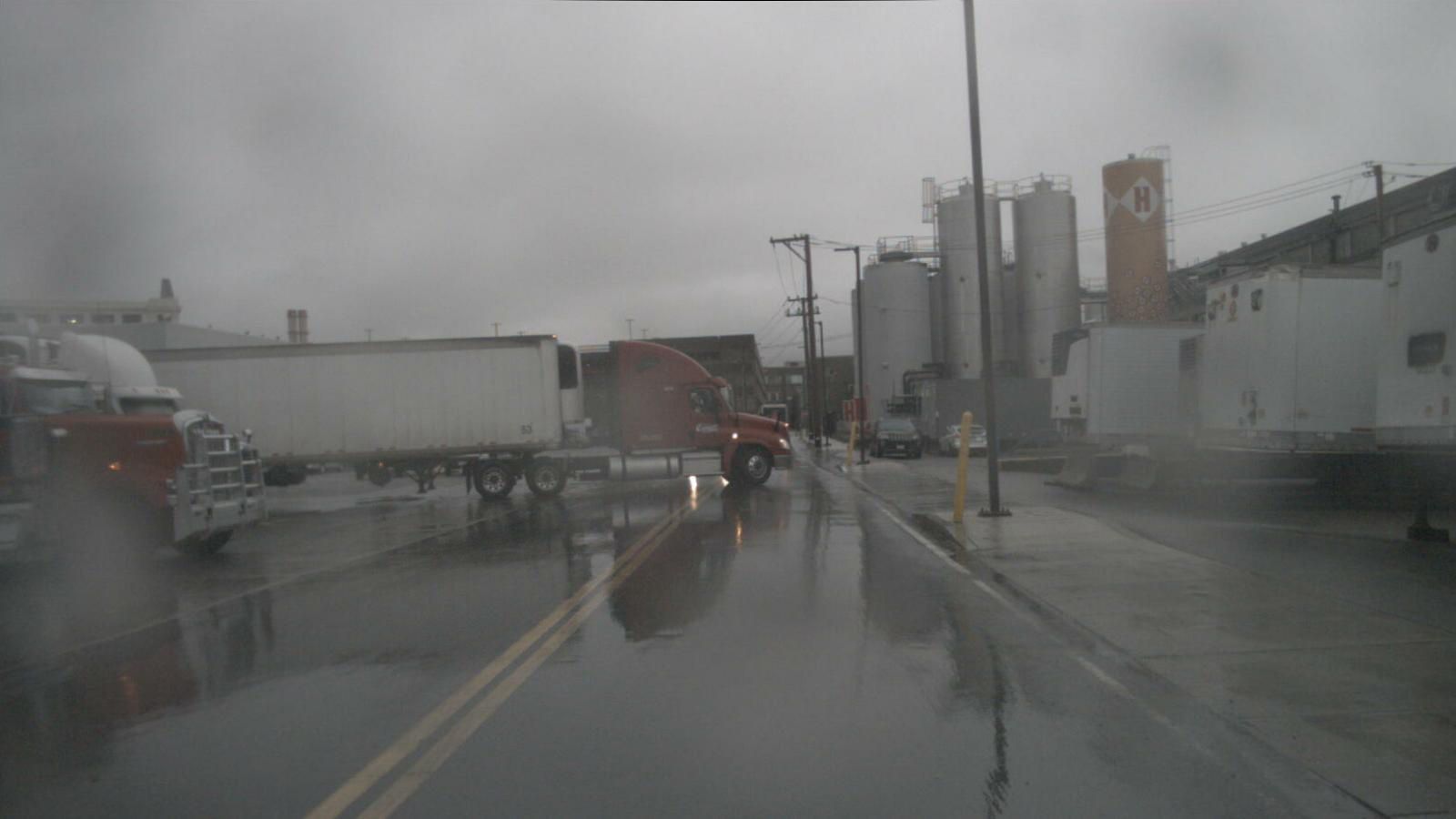}
    % \caption{Camera 1111}
  \end{subfigure}

  \vspace{0.75em}

  % --- Table ---
  \renewcommand{\arraystretch}{1.2}
  \begin{tabularx}{0.95\linewidth}{@{}lX@{}}
    \toprule
    \textbf{User} & How would the wet conditions affect the vehicle's departure from the loading zone? \\
    \midrule
    \textbf{UniVLT} & The \textcolor{ForestGreen}{wet conditions} would require the driver to be \textcolor{ForestGreen}{more cautious}, ensuring that the vehicle is properly aligned and that there is \textcolor{ForestGreen}{enough traction} before moving out of the loading zone to \textcolor{ForestGreen}{avoid any potential skidding} or loss of control. \\
    \midrule
    \textbf{Ground Truth} & The \textcolor{ForestGreen}{wet conditions} would \textcolor{ForestGreen}{necessitate a careful and measured departure}, with attention to \textcolor{ForestGreen}{maintaining traction} and \textcolor{ForestGreen}{preventing skidding}, especially when turning or braking. \\
    \bottomrule
  \end{tabularx}

  \vspace{0.75em}

  \caption{Qualitative instance of Omnidrive Object Recognition.}
  \label{fig:omnidrive-obj-recog-vis}
\end{figure}

%------ Omnidrive driving suggestion -----
\setcounter{figure}{\value{figure}}

\begin{figure}[ht]
  \centering

  % --- Images ---
  \begin{subfigure}[t]{0.32\linewidth}
    \includegraphics[width=\linewidth]{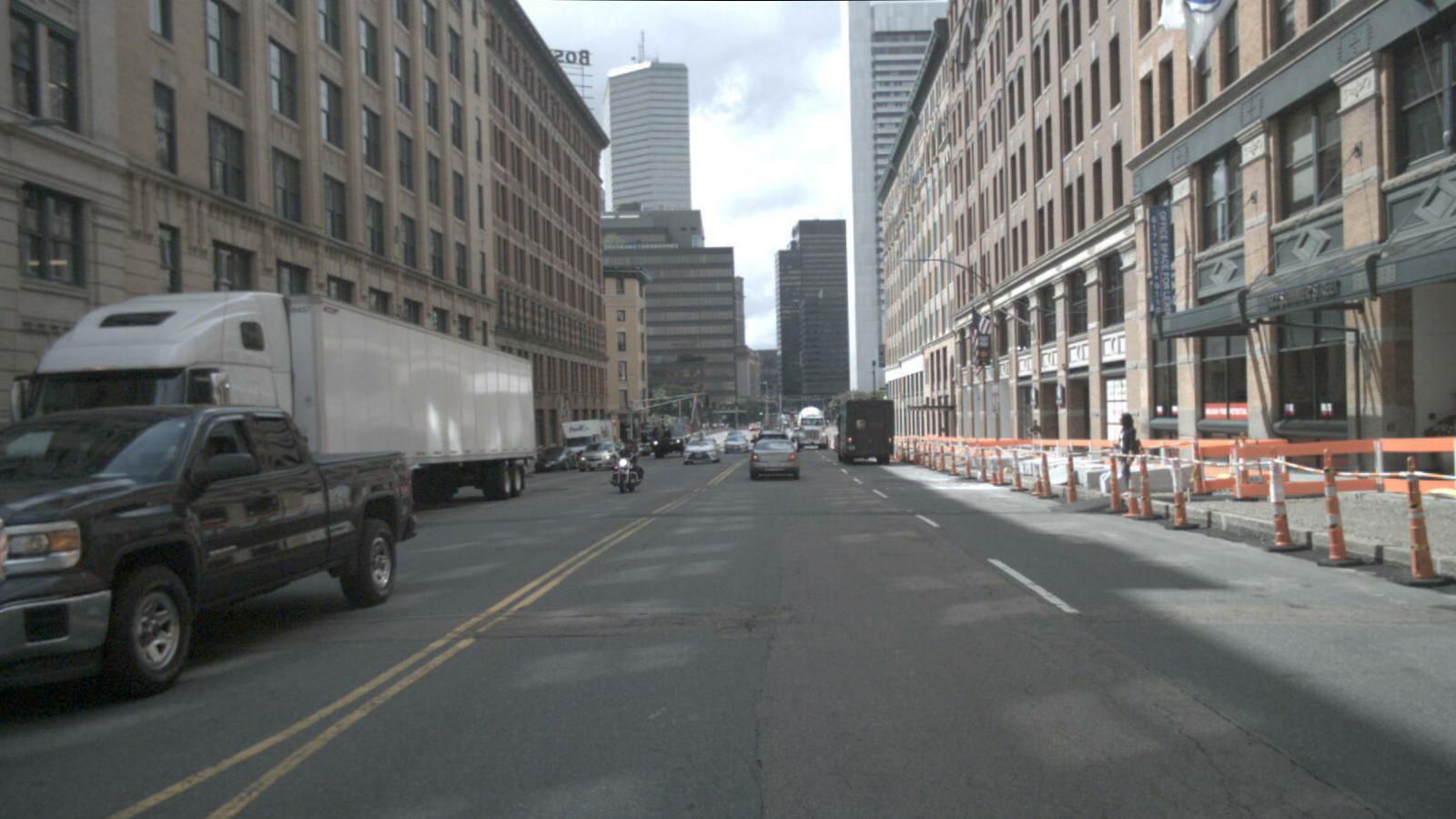}
    \caption{Frame 0}
  \end{subfigure}
  % \hfill
  \begin{subfigure}[t]{0.32\linewidth}
    \includegraphics[width=\linewidth]{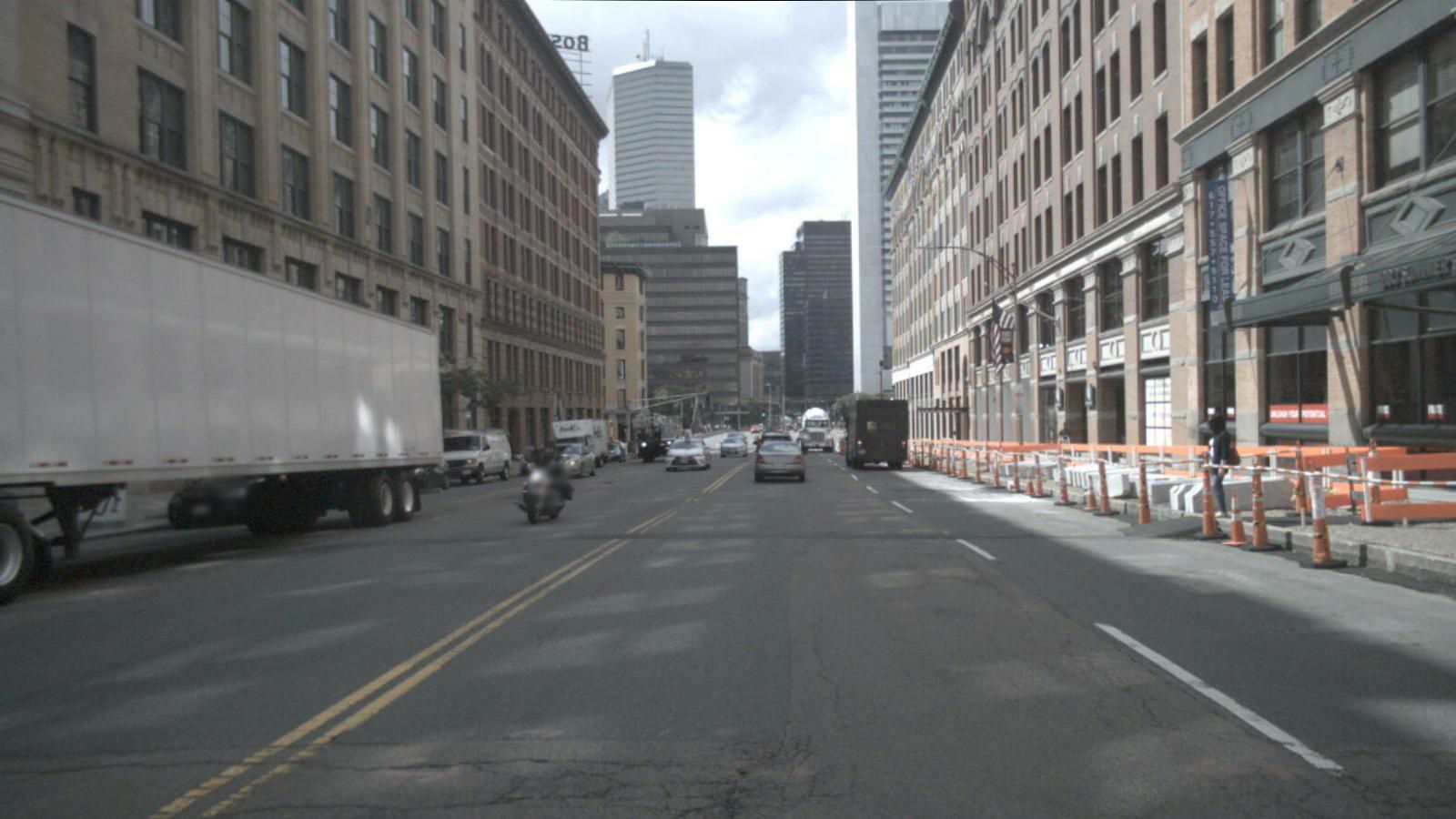}
    \caption{Frame 1}
  \end{subfigure}
  % \hfill
  \begin{subfigure}[t]{0.32\linewidth}
    \includegraphics[width=\linewidth]{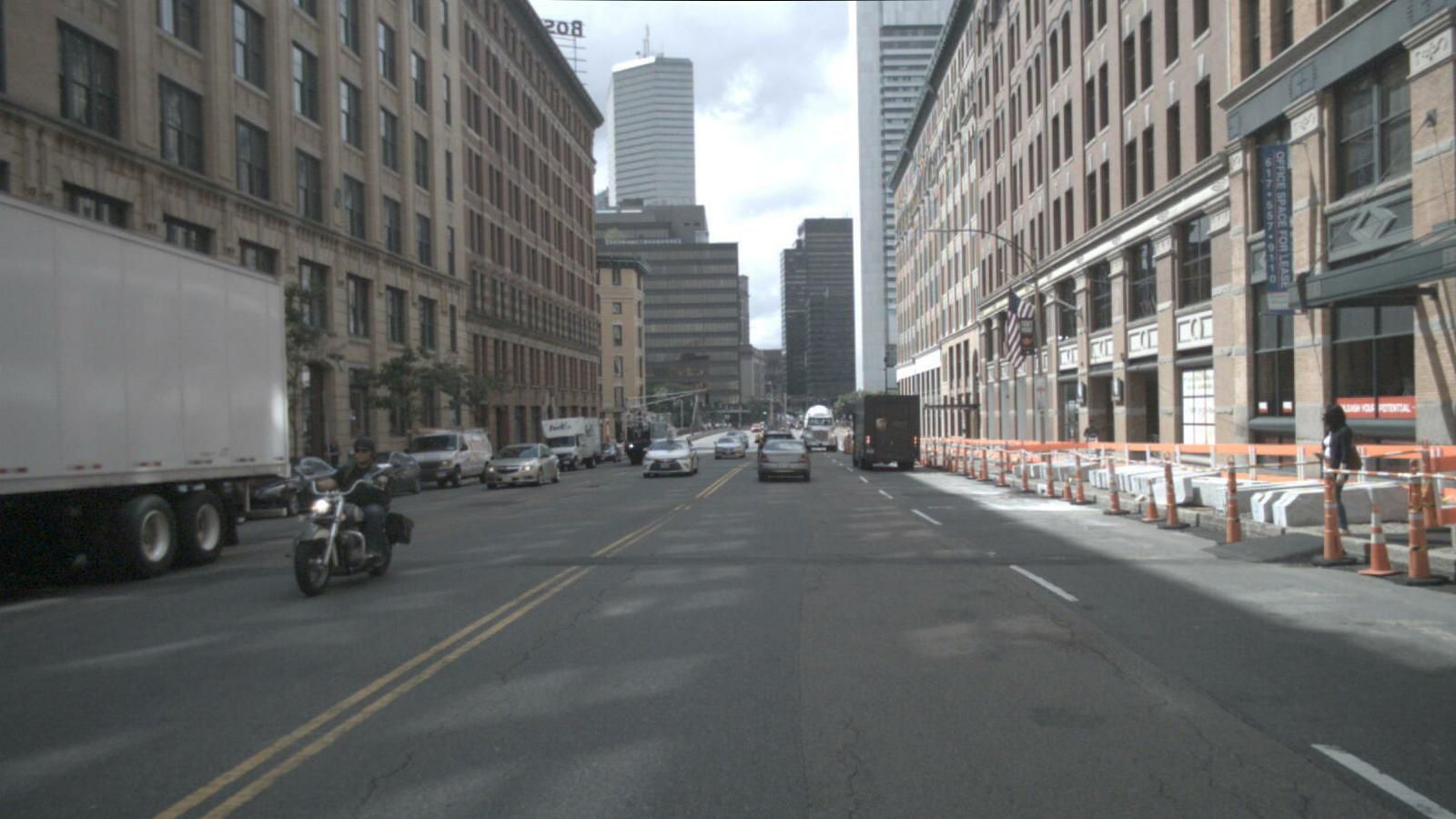}
    \caption{Frame 2}
  \end{subfigure}
  % \hfill
  \begin{subfigure}[t]{0.32\linewidth}
    \includegraphics[width=\linewidth]{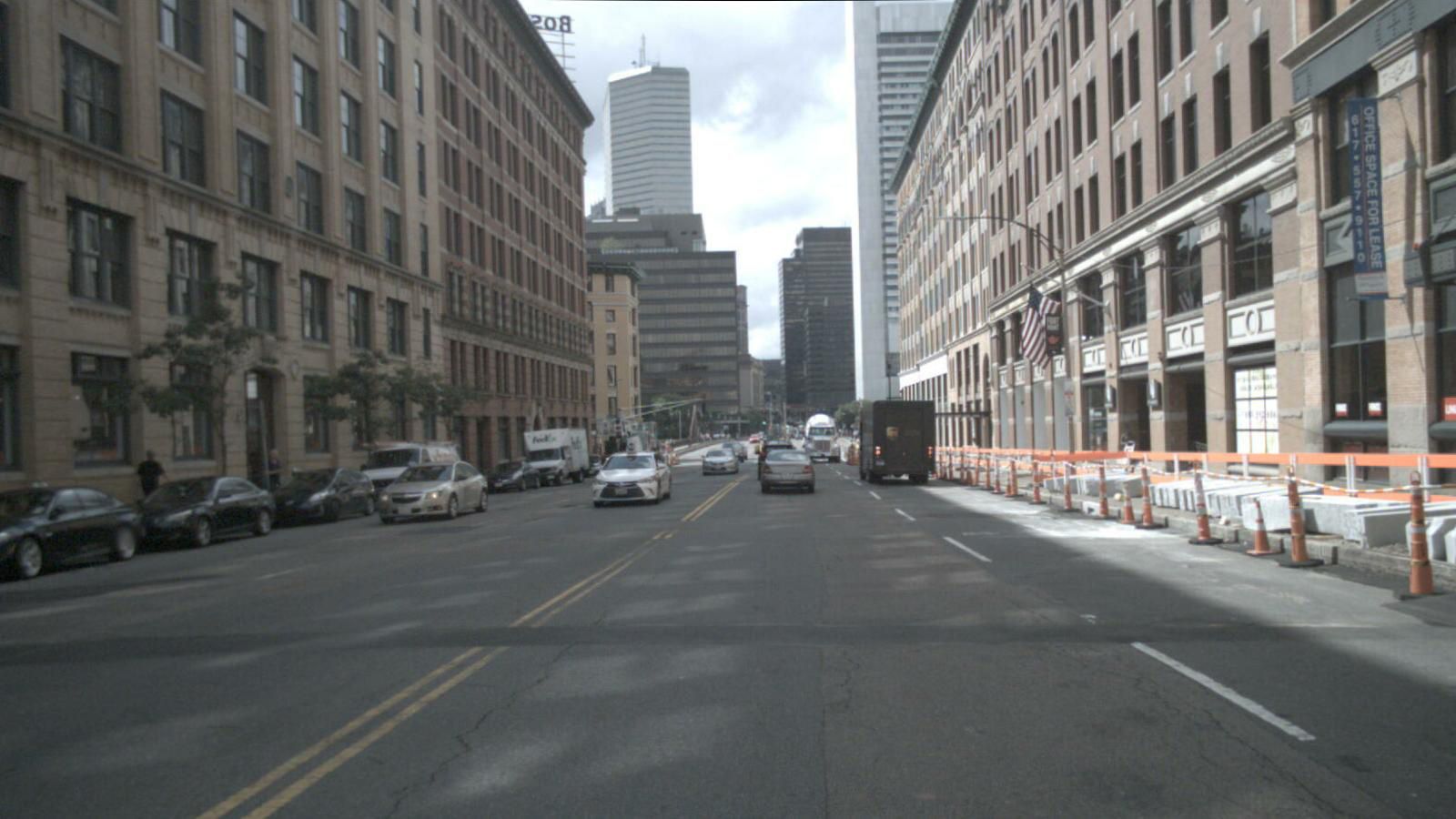}
    \caption{Frame 3}
  \end{subfigure}
  % \hfill
  \begin{subfigure}[t]{0.32\linewidth}
    \includegraphics[width=\linewidth]{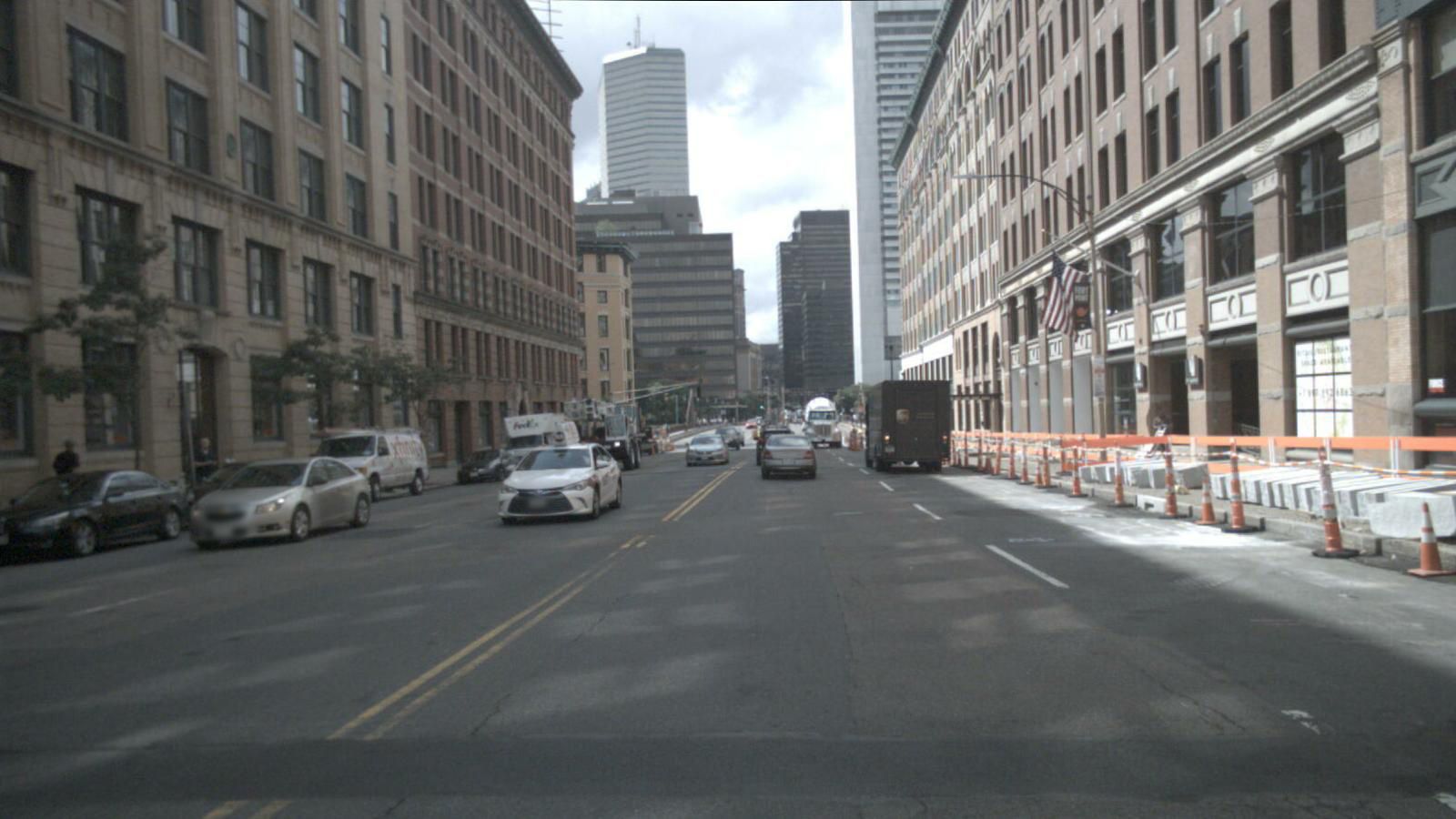}
    \caption{Frame 4}
  \end{subfigure}

  \vspace{0.75em}

  % --- Table ---
  \renewcommand{\arraystretch}{1.2}
  \begin{tabularx}{0.95\linewidth}{@{}lX@{}}
    \toprule
    \textbf{User} & What should your next action be? \\
    \midrule
    \textbf{UniVLT} & Given the current scenario, the best course of action is to \textcolor{ForestGreen}{maintain a moderate speed} and \textcolor{ForestGreen}{continue straight in our current lane}. \\
    \midrule
    \textbf{Ground Truth} & Given the current scenario, the best course of action is to \textcolor{ForestGreen}{maintain a moderate speed} and \textcolor{ForestGreen}{keep your lane while proceeding straight}. \\
    \bottomrule
  \end{tabularx}

  \vspace{0.75em}

  \caption{Qualitative instance of Omnidrive Driving Suggestion.}
  \label{fig:omnidrive-driving-sugg-vis}
\end{figure}

\end{document}